%% file: nus-thss-tmplt.tex
\documentclass[12pt,a4paper,oneside]{book}
\usepackage{url}
\usepackage{setspace}
\usepackage{float}
  \usepackage[
    backend=biber,
    style=apa,
  ]{biblatex}
\usepackage[multiple,flushmargin]{footmisc}
\usepackage{graphicx}
\graphicspath {{Imgs/}}  
\usepackage{makeidx}
\usepackage[hidelinks]{hyperref}
\usepackage{csquotes}
\usepackage{metalogo}
\usepackage{microtype}
\usepackage{caption}
\usepackage[colorinlistoftodos,textsize=tiny]{todonotes}
\usepackage[colorinlistoftodos,textsize=tiny]{todonotes}

\makeatletter
\patchcmd\@footnotetext
  {#1}
  {\toggletrue{blx@footnote}#1}
  {\togglefalse{blx@tempa}}
  {}
\makeatother

\usepackage{etoolbox}

\makeatletter
\AtEveryCitekey{%
  \ifboolexpr{ test {\iffieldequalstr{entrysubtype}{classical}}
               and not test {\iffieldundef{shorttitle}} }
    {\ifciteseen
       {\blx@ibidreset\clearname{labelname}}
       {\savefield{title}{\cbxtitle}\restorefield{labeltitle}{\cbxtitle}}}
    {}}
\makeatother

\AtEveryBibitem{\clearlist{language}}

\begin{document}
\input{./ttlpg-kcl.tex}
\input{./dclrtn.tex}
\input{./acknwldgmnts.tex}


\normalsize
\mdseries
\tableofcontents
\newpage
\input{./smmry.tex}
\listoftables
\listoffigures
\newpage
\newpage


\doublespacing
\frenchspacing

\setcounter{page}{1}
\pagenumbering{arabic}

\chapter{Introduction}
\begin{figure}
    \centering
    \centerline{\includegraphics[scale=0.4]{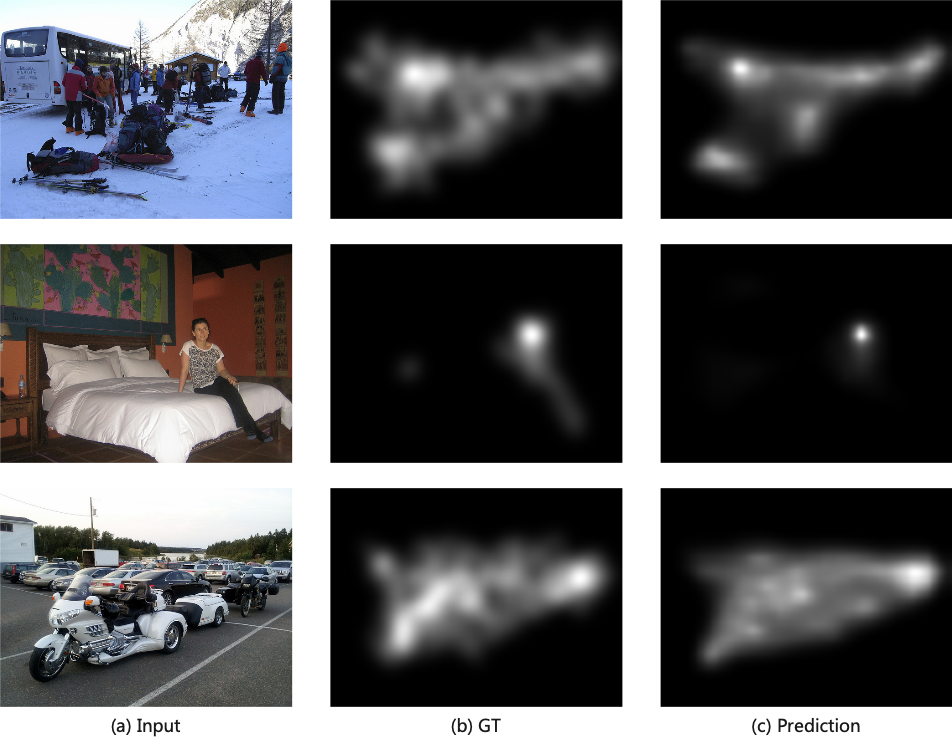}}
    \caption{Sample predictions from EML-Net \parencite{jia2020eml}, which is the state-of-the-art method on the SALICON \parencite{jiang2015salicon} dataset. The samples are released by the author.  There is still a gap between the predictions and ground truths, e.g., the attention on some objects is not well predicted.}
    \label{fig:eml}
\end{figure}

\addtocontents{toc}{\protect\enlargethispage{\baselineskip}}
Saliency Prediction has been attracting more and more researchers to enter this field, which aims to predict the human attention distribution in images. The task is defined as: given an input image, the prediction model is supposed to output the human attention distribution, i.e., the numeric saliency map. Examples are shown in Figure \ref{fig:eml}, which is the sample prediction of EML-Net \parencite{jia2020eml}. Human gaze when scanning images has been discussed for decades in cognitive science \parencite{mackworth1967gaze, lanata2013eye, fairchild2021discriminable}. The significance of modeling where people look at presents in many applications. In cognitive science, some researchers utilize saliency models for model-based hypothesis testing. In computer vision and robotics, some examples include image and video captioning \parencite{schwartz2017high}, visual question answering \parencite{lu2016hierarchical}, zero-shot image classification \parencite{jiang2017learning}, patient diagnosis \parencite{wang2015atypical}, SLAM robustness \parencite{li2020attention}, etc.
In the deep learning era, convolutional neural network based methods dominate in saliency prediction up to now, thanks to its strong image representation. Deep Gaze I \parencite{kummerer2014deep} is the first method that transfers the ImageNet \parencite{deng2009imagenet} learned features to the field of saliency. EML-Net \parencite{jia2020eml} transfers the pretrained features from NasNet \parencite{zoph2018learning}, a very large convolutional network, and achieves competitive performance on both SALICON \parencite{jiang2015salicon} and MIT300 \parencite{Judd_2012} benchmarks. CASNet \parencite{fan2018emotional} proceeds to exploit the relation between human saliency and the image sentiment based on VGG \parencite{simonyan2014very} net. All these work shows the effectiveness of transfer learning and the possibility of human saliency prediction. In the most recent years, more advanced machine learning methods are proposed and the performance of the saliency model progresses a lot. Deep Gaze IIE \parencite{linardos2021deepgaze} utilizes model ensembling, where it effectively combines twelve backbones, and pulls up the score on the MIT300 benchmark \parencite{Judd_2012}. UNISAL \parencite{droste2020unified} considers the domain shift between the static image and dynamic video saliency. However, the evaluation scores of the recent state-of-the-art methods on MIT300 benchmark \parencite{Judd_2012} are still far from the gold standard shown in Table \ref{table:relatedwork}. The gold standard means a Gaussian kernel density estimate which stands for the average saliency distribution of the participants in the process of building the dataset. We can find in Table \ref{table:relatedwork} that there is still a gap between the state-of-the-art and the human standard. Also, on the SALICON benchmark \parencite{jiang2015salicon}, the difference between the current prediction and the ground truth is still obvious. Examples are shown in Figure \ref{fig:eml}.  The problems of the gap are mainly two: a) the image representation and model capacity of the recent state-of-the-art methods on the MIT300 benchmark \parencite{Judd_2012} are not strong enough, due to the limitation of the convolutional kernel, b) most of the recent methods ignore the benefits of high-level factors that affect our gaze control. More details are demonstrated as follows.

\begin{figure}
    \centering
    \centerline{\includegraphics[scale=0.6]{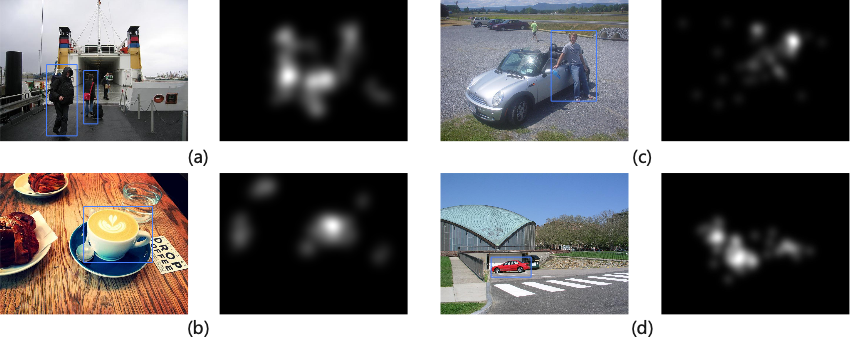}}
    \caption{The objects in a scene are highly correlated with human attention. On the left are the original inputs and the human saliency annotations are shown on the right pictures, which are selected from SALICON dataset \parencite{jiang2015salicon}. People tend to allocate most of their attention to some common objects.}
    \label{fig:intro}
\end{figure}

To this day, much research focuses on the traditional convolutional network but few makes use of the recent popular self-attention structure, Transformer \parencite{vaswani2017attention}. The Transformer is first proposed for NLP tasks, e.g., machine translation, and consists of an encoder and a decoder which share a similar structure. ViT \parencite{dosovitskiy2020image} introduces the Transformer encoder layer into image classification and now it has outperformed the traditional CNN on ImageNet \parencite{deng2009imagenet}, which establishes the potential of the Transformer in computer vision. The improved image representation and also the global cues gained from the Transformer encoder are still not utilized in the field of saliency. Inspired by the human perception process that people tend to get a global view of a picture when allocating attention to its various regions, we employ the Transformer encoder inside our model to enhance the global cue of the input image. The bi-directional self-attention in the Transformer could extract a more comprehensive feature, which we show by visualization in the experiment. The global feature from the Transformer is also preserved through the skip connection in the decoder network. 

On the other hand, many recently proposed methods mainly employ advanced machine learning or deep learning techniques, e.g., GAN \parencite{goodfellow2014generative}, LSTM \parencite{hochreiter1997long}, domain adaptation \parencite{wang2018deep}, etc., but few research efforts attempt to take advantage of the realistic human perception process. When people look at an image, we tend to recognize the common objects in the scene, which is shown in Figure \ref{fig:intro}, and meanwhile determine which areas are salient. Inspired by this, simultaneous semantic segmentation is introduced in our method to simulate the human eyes. \parencite{cerf2007predicting, schauerte2012predicting} propose to utilize a face detector to determine the visual saliency. \parencite{cordel2019emotion} utilizes Mask RCNN \parencite{he2017mask} for emotional detection to enhance the saliency prediction. In \parencite{fan2019s4net}, object detection is utilized for saliency detection. The objects are first localized and then determined whether it is salient. The methods utilizing object detection mainly transfer the knowledge of object perception in the second stage reasoning of the detector or directly take advantage of the detection results. However, object detection cannot give comprehensive information as semantic segmentation. Due to the region proposal \parencite{ren2015faster} process, the recent object detection networks mainly focus on the features inside the proposed bounding boxes in the second stage reasoning on an object level while semantic segmentation can more explicitly model the object and the background information on a pixel level simultaneously. In addition, the information of the objects not captured by the semantic segmentation is reserved in the background feature, which is lost by the object detection in its second stage or the detection results. Hence in this paper, we propose to utilize semantic segmentation to enhance our saliency result as a new exploration.

First, we would give a brief overview of this thesis. Chapter 1 gives the introduction of the field of saliency prediction and some problems of the recent research efforts. In Chapter 2, we would give a  summary of the related work, including saliency prediction, Transformers, and multi-task learning in saliency prediction and detection. In Chapter 3, we first show our investigation result on the human gaze control in the real scenes about different types of human gaze control and different factors that may affect where we look. Then we introduce the principle of our Transformer-based model which simultaneously completes the object segmentation. Followed by Chapter 4, our experiment shows the effectiveness of the Transformer and that the closely correlated multi-task learning greatly benefits the main task performance. Our method achieves competitive performance against the recent state-of-the-art models. In Chapter 5, we give a summary of this thesis and discuss about the future work of our method.

Our main contributions are as follows.\\
\indent1. We employ the Transformer encoder inside our saliency model, which enhances the image representation by incorporating the global cues from the input image. The image representation gained from the Transformer is much stronger than the traditional CNN which is commonly utilized in the recent methods. Transformer gains a performance lift from 0.765 to 0.772 in AUC-Judd and from 0.651 to 0.655 in s-AUC on SALICON validation set. We are the first to introduce the hybrid Transformer encoder in the field of saliency.\\
\indent2. We introduce semantic segmentation into saliency prediction, which simulates the human perception process and improves the performance of the saliency model. Few recent methods consider the high-level factors that control our gaze, which can lead to a performance leap, and semantic segmentation could better model the object and the background information on a pixel level simultaneously and comprehensively than recent object detection methods which focus on the features inside the proposed bounding boxes \parencite{ren2015faster} in the second stage reasoning. To the best of our knowledge, we make the first attempt to model the saliency and semantic segmentation simultaneously. The joint learning gains a performance lift from 0.772 to 0.774 in AUC-Judd and from 0.655 to 0.656 in s-AUC on SALICON validation set.\\

\newpage
\chapter{Related Work}
\addtocontents{toc}{\protect\enlargethispage{\baselineskip}}
Saliency prediction has been researched for more than a decade. As deep learning started showing its potential, researchers in this field start using more complex neural networks to solve the problem, because of the enhanced representation of images. Some researchers employ sophisticated methods for further improvement, e.g. multi-task learning, model ensembling, etc. In this chapter, we would provide a literature review of saliency prediction and multi-task learning in saliency detection and prediction. In addition, while Transformers become more popular, relatively few research efforts related to Transformers are done for saliency prediction. So we propose to introduce this powerful global representation for the saliency prediction and in this chapter, we would review Transformers.

\section{Saliency Prediction}

Deep Gaze I \parencite{kummerer2014deep} extracts feature maps from all levels and then does filtering followed by softmax, which results in a probabilistic output. Deep Visual Attention \parencite{wang2017deep} takes the feature maps from the last three layers in vgg \parencite{simonyan2014very} as the input of its decoder where deconvolution is used to recover the resolution to get three saliency maps which are then fused to get the final result. The network takes advantage of the powerful representation of vgg net and demonstrates multi-level feature fusion could be useful. Saliency Attentive Model \parencite{cornia2018predicting} introduces attentive LSTM to progressively modify and enhance the feature maps for saliency prediction. Learnable center priors are also introduced to simulate the human eye. In SalGAN \parencite{pan2017salgan}, the adversarial method is introduced to make the output resemble the ground truth. FES \parencite{rezazadegan2011fast} estimates the human saliency by Bayesian network and could be used in real-time. LDS \parencite{fang2016learning} explores different subspaces which have different capabilities for distinguishing targets and distractors. CovSal \parencite{erdem2013visual} introduces region covariance to integrate different feature dimensions. eDN \parencite{vig2014large} follows an entirely
automatic data-driven approach that searches for optimal features. \parencite{itti1998model} combines multi-scale image features and utilizes a dynamic neural network to select attended
locations in order of decreasing saliency. In CASNet \parencite{fan2018emotional}, emotional factor in saliency is considered and the model is designed to recognize the emotion of different objects. GazeGAN \parencite{che2019gaze} proposes an adversarial training method that assigns two discriminators to coarse-scale and fine-scale outputs. The author also proposes effective data augmentation for saliency prediction. TranSalNet \parencite{lou2021transalnet} is among the few research efforts which introduces the Transformer into the field. The network adds three Transformer blocks onto the last three layers of the backbone, to enhance the global features. Deep Gaze II \parencite{kummerer2017understanding}, based on Deep Gaze I, proposes a readout network to better transfer the features from the pretrained model to saliency knowledge. MSI-Net \parencite{kroner2020contextual} proposes an  encoder-decoder structure where a dilation convolution module to increase the receptive field is between them. This could also inspire us that global features are important cues for saliency prediction. EML-Net \parencite{jia2020eml} uses transfer learning to gain a stronger representation of input images. Unisal \parencite{droste2020unified} proposes multi-task learning, i.e. unified learning of the static image saliency prediction and dynamic video saliency prediction, to enhance both tasks. Deep Gaze IIE \parencite{linardos2021deepgaze} is the ensemble version of Deep Gaze, which combines more than ten models. The author talks about the effectiveness of different backbones and different ways of model ensembling. DI-Net \parencite{yang2019dilated} proposes Dilated Residual Network and Dilated Inception Module to enhance the receptive field. CEDNS \parencite{qi2019convolutional} is a network
taking advantage of the structure of the DenseNet \parencite{iandola2014densenet}, which takes advantage of feature connections to enhance the image representation. Table \ref{table:relatedwork} shows the summary of the related work in saliency prediction.

\begin{table}[h!]
\centering
\caption{Related work in saliency prediction on the MIT300 benchmark \parencite{Judd_2012}. The benchmark provides two types of evaluation determined by the model submitted: probabilistic and traditional evaluation. \textbf{Note that probabilistic evaluation regards the result as a mathematical distribution, so the evaluation will compute the best result saliency map for each metric, which leads to a higher score in AUC than traditional evaluation.}}.
\begin{tabular}{c c c c c} 
 \hline
 Method & Year & Type of evaluation &Type of method&AUC \\ 
 \hline\hline
 Gold Standard&-&probabilistic&-&0.9341\\
 DeepGaze IIE&2021&probabilistic&model ensembling&0.8829\\
 UNISAL&2020&probabilistic&domain adaptation&0.8772\\
 EML-NET&2018&traditional&model ensembling&0.8762\\
 MSI-Net&2020&probabilistic&CNN&0.8738\\
 Deep Gaze II&2017&probabilistic&transfer learning&0.8733\\
 TransalNet&2021&traditional&Transformer model&0.8730\\
 GazeGAN&2019&traditional&GAN model&0.8607\\
 CASNet&2018&traditional&CNN&0.8552\\
 SAM-Resnet&2018&traditional&LSTM model&0.8526\\
 SalGAN&2017&traditional&GAN model&0.8498\\
 DVA&2018&traditional&CNN&0.8430\\
 Deep Gaze I&2015&probabilistic&transfer learning&0.8427\\
 eDN&2014&traditional&classic model&0.8171\\
 CovSal&2013&traditional&classic model&0.8116\\
 LDS&2016&traditional&classic model&0.8108\\
 FES&2011&traditional&classic model&0.8018\\

 \hline
\end{tabular}
\label{table:relatedwork}
\end{table}

In saliency prediction, much research has been done to improve the ability of image representation using CNN, e.g. Deep IIE, EML-Net, DVA, etc. But few research efforts focus on how global features affect the saliency result. In this thesis, we try to explore the effectiveness of including global features from the Transformer. Some research focuses on introducing multi-task training to enhance the saliency prediction. Here we propose another task, semantic segmentation, which could improve the training of saliency.

\section{Transformers}
The Transformer is first proposed in \parencite{vaswani2017attention}, which is initially designed for NLP tasks. It achieves state-of-the-art performance in many NLP tasks, which demonstrates the effectiveness of large-scale pretraining. Bidirectional multi-head self-attention is the key idea of Transformer which could give a more comprehensive representation of a sequence than traditional LSTM. LSTM could not handle a very long sequence, but the Transformer could. In recent years, the transformer has come into the computer vision field. ViT \parencite{dosovitskiy2020image} is the first vision Transformer that regards an image as "words", taking advantage of the Transformer encoder block to get a powerful global image representation. BoTNet \parencite{srinivas2021bottleneck} proposes to insert the multi-head self-attention module into Resnets, where one position in the feature map is seen as a single patch. DS-Net \parencite{mao2021dual} explores the representation capacity of local and global features for image classification. Conformer \parencite{peng2021conformer} fuses local features and global representations under
different resolutions in an interactive fashion. TransUnet \parencite{chen2021transunet} employs the Resnet plus the Transformer encoder to enhance the feature for medical image segmentation.

Many efforts and experiments demonstrate the effectiveness of the image representation of Transformers which is based on the complete bidirectional self-attention and successfully captures a global view of the image. In our work, the global view of an image is consistent with the human perception process, where we first run down an image and then "decide" where we look. Inspired by this, we design a model based on the image representation of the Transformer.

\section{Multi-task Learning in Saliency Prediction and Detection}
Specific to saliency prediction and detection, some researchers propose to introduce multi-task learning to enhance the performance of either the main task or multiple tasks, e.g., salient object subitizing, camouflaged objects segmentation. TINet \parencite{zhu2021inferring} demonstrates the boundary-aware salient object detection could be more robust in various scenes and datasets since boundary detection could be seen as texture guidance for salient object detection. In \parencite{Kruthiventi_2016_CVPR}, the joint learning of human fixation prediction and salient object detection could enhance both tasks simultaneously, due to the close correlation between eye fixation and object saliency. \parencite{Chen_2017_ICCV} demonstrates that the salient objects in a complex scene could be detected by pre-segmented semantic objects that receive the highest human attention. In \parencite{He_2017_ICCV}, the joint learning of salient object subitizing and detection is proposed, where subitizing could be strong guidance for the main task, salient object detection. Parameters in an adaptive weight layer of the salient object detection task is dynamically determined by an auxiliary subitizing network. ASNet \parencite{Wang_2018_CVPR} predicts human attention and salient object segmentation at the same time and convolutional LSTM in the network is used to refine the feature in each stage. \parencite{Islam_2018_CVPR} proposes simultaneous learning of detection, ranking, and subitizing of multiple salient objects. \parencite{Li_2021_CVPR} proposes unified learning of salient object detection and camouflaged object detection since the camouflaged objects in a scene tend to be salient from human perception.\\

\noindent Inspired by the recent efforts on Transformers and multi-task learning in the field of saliency, we would introduce our Transformer-based model which simultaneously performs the semantic object segmentation in the following chapter. Also, we would provide human study evidence to bridge the gap between deep learning based saliency prediction and cognitive science understanding of human eye control.

\chapter{Proposed Method}
\addtocontents{toc}{\protect\enlargethispage{\baselineskip}}
\section{Human Gaze Control in the Real World}
In cognitive science,  how gaze control operates over complex real-world scenes has become a central concern in several cognitive science related disciplines including cognitive psychology,
visual neuroscience, and machine vision. Predicting the human eye fixation with deep learning could also be regarded as a way of simulating the human gaze. In this chapter, we would review cognitive studies concerning human visual perception and explain the motivation of our model. 

\subsection{Types of Human Gaze Control}
The human gaze control could be generally divided into two parts: stimulus-based gaze control and knowledge-driven gaze control \parencite{henderson2003human}. Stimulus-based gaze control refers to the bottom-up stimulus-based information generated from the image. Humans tend to fixate on a certain patch of a scene due to its certain distinguished properties which make it salient, e.g.,  high spatial frequency
content and edge density \parencite{mannan1996relationship, mannan1997fixation}, color, intensity, contrast, edge orientation, etc. Knowledge-driven gaze control refers to top-down memory based knowledge generated from internal visual and
cognitive systems, including meaning derived from the previously fixated \parencite{henderson1999effects} objects and the goals
and intents of the viewer \parencite{henderson2003human}. Also, the global view and spatial layout from the first fixation provide important information about where the interesting objects could be \parencite{oliva2003top}. In our model, we employ the Transformer encoder to enhance the global feature, where bi-directional self-attention models the relation between patches of a scene. The deepest layer feature of the Transformer simulates the human knowledge-driven gaze control after the first fixation.

\subsection{What Affects Where We Look?}
What affects where we look has become a significant problem of interest since scientists were able to capture the human eye movement. Mainly four factors simultaneously contribute to our eye fixation: salience, object recognition, value, and plans for saccadic target selection \parencite{schutz2011eye}. Similar to Section 3.1, salience refers to distinguished visual attributes of a region, e.g., intensity, color, and orientation. When looking around, the world is full of objects, so it would be a natural assumption that object recognition motivates human eye movement. \parencite{einhauser2008objects} demonstrates that objects predicted gaze performs better than salience (features) predicted  gaze. \parencite{morvan2010observers} supports the role of object-based saccadic target selection. \parencite{cerf2009faces} finds that people tend to look at faces or texts. Value and Plans could be categorized into knowledge-driven gaze control. Value refers to the consequences, e.g., the reward gained after looking somewhere, taken into account when human eyes are selecting targets and plans refer to a certain task or motion directing one's fixations \parencite{schutz2011eye}. Our model introduces object segmentation to simulate the "object recognition" factor affecting our eye attention. Through the multi-task learning, the model would be benefited from object perception.

\section{Architecture Overview}
\begin{figure}
    \centering
    \centerline{\includegraphics[scale=0.6]{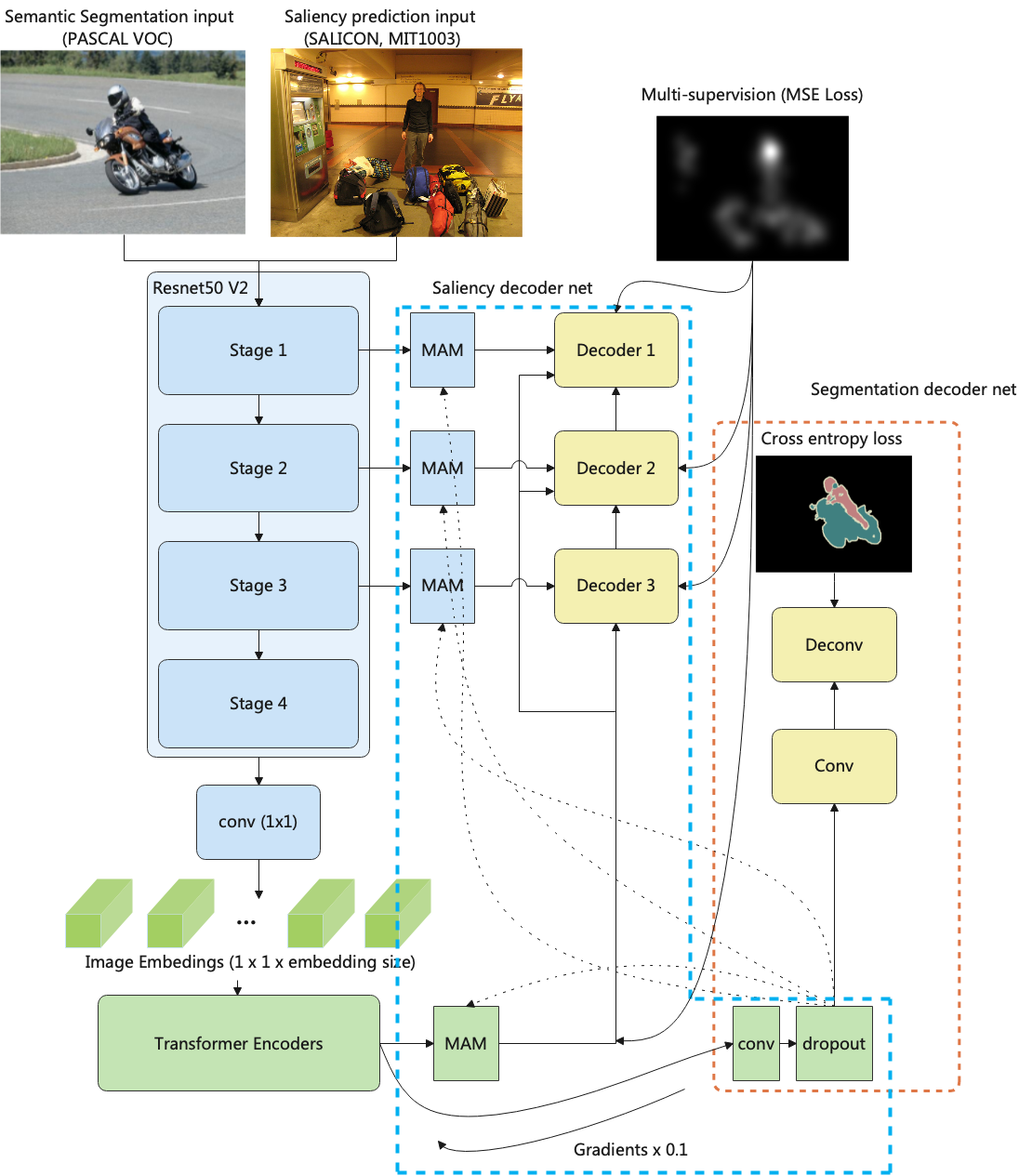}}
    \caption{Architecture overview. The Transformer and Resnet serve as the image encoder. Blue and red boxes represent the decoders of saliency prediction and object segmentation, respectively. Multi-task attention module (MAM) introduces features from semantic segmentation for feature selection. The knowledge from segmentation can be transferred because most common objects are included in the segmentation dataset.}
    \label{fig:archi}
\end{figure}

In this chapter, we will discuss the computational model we design for saliency prediction. Our model is called \textbf{SSETM} (\textbf{S}emantic \textbf{S}egmentation \textbf{E}nhanced
\textbf{T}ransformer \textbf{M}odel for Human
Attention Prediction). The main part of the network is the bottom-up structure, which combines both local features from traditional CNN and the global feature gained from self-attention in the Transformer, shown in Figure \ref{fig:archi}. The bottom-up structure effectively fuses features from different stages and the skip connection could make the global feature better propagate across each fusion stage. The sub-branch of the network is for semantic segmentation. Since there are few datasets containing both saliency and semantic segmentation ground truths, we train this branch upon the Pascal VOC 2012 dataset which includes common object categories and transfer the knowledge into the main branch using a 1$\times$1 convolution kernel. The feature for segmentation is merged into saliency prediction, which intrinsically enhances the main task. This could simulate the human perception process, i.e. people recognize objects given an image and assign attention at the same time.

\section{Transformer Encoder}
\begin{figure}
    \centering
    \centerline{\includegraphics[scale=0.5]{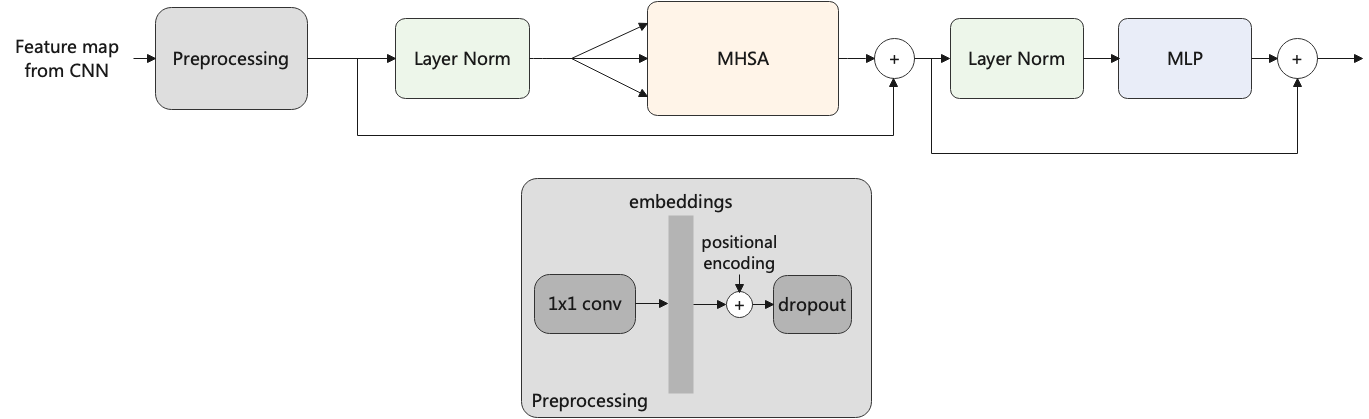}}
    \caption{The Transformer encoder. MHSA refers to the multi-head attention module.}
    \label{fig:trans}
\end{figure}

As human eyes tend to get an intrinsic global view when allocating attention onto an image \parencite{oliva2003top}, we consider enhancing the global feature from the traditional CNN. To further enhance the global cues, one way is by enlarging the receptive field of the feature map, which is often implemented by dilated convolution \parencite{yu2015multi}, but self-attention in the Transformer regards the image as a sequence and computes the mutual effects between each pair of the regions in the image. Each region in an image would attend to the whole picture, so this would introduce more global cues than the traditional transformed convolution which is constricted by its kernel size. The self-attention computing process in the Transformer could also explain the human eye fixation and saccades since we always intrinsically glance at different regions in a few milli-seconds given an image and then determine which areas are salient. The Transformer serves as part of the encoder network for both saliency prediction and semantic segmentation. The encoder consists of Resnet and vision Transformer, shown in Figure \ref{fig:archi}. We use Resnet50 V2 (pre-activation) as the backbone to extract deep representations of input images including feature maps of four levels. The deepest feature map has the largest receptive field and more global feature, and its size is also the smallest, so it could be sent into the Transformer block to enhance the global representation, considering both aspects of effectiveness and efficiency. The sizes of feature maps of other levels are too large, which would require too many computing resources. The Transformer is proposed to deal with sequence inputs so one pixel in the deepest feature map from Resnet50 is treated as one patch for the vision Transformer and the whole feature map is treated as a sequence with a length of H$\times$W (H, W stand for the height and width of the feature map, respectively), where each patch corresponds to a single region of the input image. 1$\times$1 convolution is implemented before Transformer modules to change the feature to the embedding size of the Transformer. The positional embedding is initialized from pretrained ImageNet model. 2D Dropout follows to decrease overfitting. The principle of the Transformer encoder is shown in Figure \ref{fig:trans}. The process before the Transformer encoder could be formulated as the following:
\begin{equation}
    fe = R(i)
\end{equation}
\begin{equation}
    emb^* = Flatten(E(fe))
\end{equation}
\begin{equation}
    emb = Dropout(emb^*  + p)
\end{equation}
\emph{i} refers to the input, \emph{R} refers to Resnet, \emph{fe} refers to the extracted feature map, \emph{E} refers to the embedding layer, and \emph{p} refers to the positional encoding before Transformer encoders. The shape of feature maps after Resnet is \emph{(B, C, H ,W)}. The embedding layer consists of a 1x1 convolution kernel with the stride 1 and output channels of the embedding size of the Transformer. The shape of the result after the embedding layer is \emph{(B, Embedding Size, H, W)}. The shape of \emph{embs*} in Equation 3.2 is \emph{(B, H*W, Embedding Size)}.

Same as \parencite{vaswani2017attention}, the image sequence embeddings pass through multiple Transformer encoders to enhance the global features. Transformer encoders consist of a multi-head self-attention (MHSA) module and MLP, in order. MHSA is divided into multiple heads to learn different knowledge from different feature spaces \parencite{vaswani2017attention}. The bidirectional self-attention regards the image as a sequence and computes attention weights between each pair of patches, helping the model learn more global cues. The computing process of one Transformer encoder could be summarized as the following:
\begin{equation}
    e^* = MHSA(LN(e)) + e
\end{equation}
\begin{equation}
    o = MLP(LN(e^*)) + e^*
\end{equation}
\emph{e} refers to image embeddings. LN refers to layer normalization. \emph{o} refers to the output features.   
The image embeddings would pass through multiple Transformer encoders. The final layer normalization change the feature distribution to a proper position:
\begin{equation}
    o^* = TEs(o)
\end{equation}
\begin{equation}
    f = LN(o^*)
\end{equation}
\emph{TEs} refer to multiple encoders. \emph{o} is the same as in Equation 3.5. \emph{f} refers to the output global feature.

\section{Multi-task Attention Module}
\begin{figure}
    \centering
    \centerline{\includegraphics[scale=0.6]{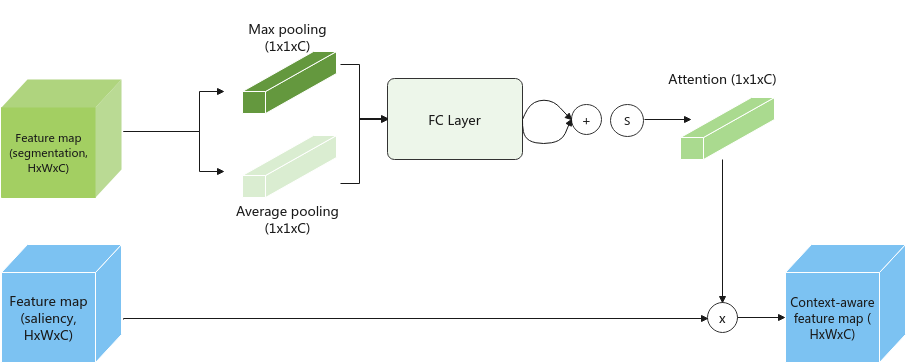}}
    \caption{Multi-task attention module (MAM). "S" refers to sigmoid activation. Similar to \parencite{woo2018cbam}, the final attention assigns a weight to each channel of the feature map from the main task. }
    \label{fig:cb}
\end{figure}

Multi-task Attention Module (MAM), shown in Figure \ref{fig:cb}, deals with the interaction between multiple tasks. In  practice, we find that simply adding another subtask onto the network architecture of the main task is not very useful and effective. That is because another subtask would always share some weights with the main task. In our network, the shared part is the encoder. The encoder has to learn different aspects of the input image at the same time to get the best result for each task. The feature representation for the input needed for each task could not be the same, so one of the problems of introducing multiple tasks is how to fairly deal with the interaction between tasks. Therefore we design the MAM module to merge the feature from the subtask into the network of the main task, as attention weights or feature selection. In Figure \ref{fig:archi}, we take the feature from the subtask and transfer the knowledge through one 1$\times$1 convolution. Then the transferred feature map is sent into the cross-task attention module. Similar to \parencite{woo2018cbam}, max pooling and average pooling are utilized to transform the feature map into vectors. MLP is utilized to compute the attention weight, which determines the importance assigned to each channel in the feature map of the main task. The summary is shown as the following:
\begin{equation}
    att^* = MLP(MP(s)) + MLP(AP(s))
\end{equation}
\begin{equation}
    att = Sigmoid(att^*)
\end{equation}
\begin{equation}
    fa = f^* * att
\end{equation}
\emph{s} stands for the transferred feature from the semantic segmentation network. \emph{MP} and \emph{AP} refer to max pooling and average pooling, respectively. The weight of the \emph{MLP} is shared. \emph{f*} refers to the features from the encoder and \emph{fa} refers to the feature whose channels are assigned attention weight.

Each channel in a feature map reflects one aspect of the input image, i.e. each channel could be seen as an attribute of the input image \parencite{woo2018cbam}. The attention module could automatically do the proper feature selection. In our case, the feature in the semantic segmentation network, which contains knowledge or perception of the scene and objects, helps the saliency prediction network learn the proper aspects of the input. This resembles our human visual perception process.

\section{Saliency Decoder Network}
\begin{figure}
    \centering
    \centerline{\includegraphics[scale=0.6]{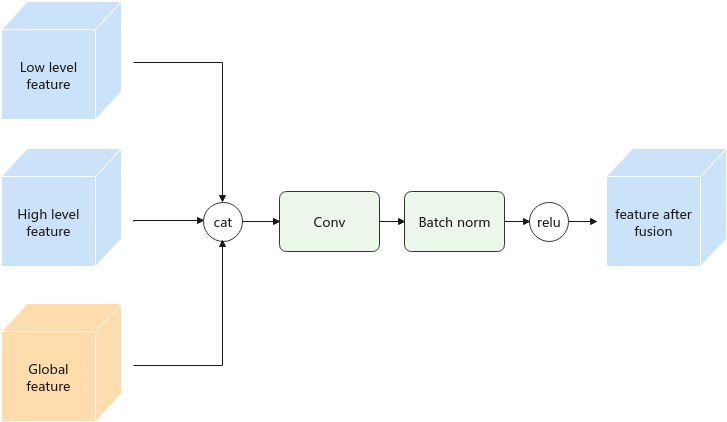}}
    \caption{Saliency subnet decoder.}
    \label{fig:my_label}
\end{figure}

The architecture of the decoder part is a bottom-up feature fusion structure, which is shown in \ref{fig:my_label}. It combines local features from CNN and global cues from the Transformer by upsampling and concatenation. Low-level feature tends to focus on detailed information of the input while the high-level feature tends to focus on context information, so combining them is needed for better information integration. Also, we repeatedly utilize the global feature from the Transformer to make it propagate across the decoder network, which is effective to retain the global cues. Here to further enhance the fusion, we utilize multi-supervision as guidance for each decoder module. Only computing the loss function upon the final output cannot guarantee other modules learn task-specific information and knowledge. So we get a result from each decoder module and resize it to the ground truth. Under the supervision of each stage of the decoder, we could make the decoder learn more specific information of the saliency prediction task. 

\section{Unified Learning of Semantic Segmentation}
\subsection{Semantic Segmentation Subnet}
How human eyes assign attention to a given picture depends on the salient extent that the objects and background have. People tend to recognize the objects and the background when they tell what attracts their attention. Inspired by this, we add a subnet for semantic segmentation and make the whole network be able to perceive objects and background. We do not employ object detection because they mainly focus on the features inside certain proposed bounding boxes, discussed in the introduction section.  In section 3.3, we design the block to transform the semantic segmentation feature for feature selection in the saliency prediction network. In Figure \ref{fig:archi}, the two tasks share the same weight of the Transformer encoder. This could also make the encoder able to learn from the feature space which is useful for scene perception. Because there are few datasets including both saliency and semantic segmentation ground truths, and to make our work more convenient for practical use, e.g. we may train the model upon many different saliency prediction datasets, we choose to train the segmentation subnet on the Pascal VOC 2012 dataset, which contains the common objects most of which present in the saliency prediction dataset as well, e.g. people, cars, motorcycles, etc. This helps transfer the knowledge from object segmentation into saliency prediction. The unified learning is across datasets, which is also employed in OMNIVORE \parencite{girdhar2022omnivore}. OMNIVORE utilizes a single model to train jointly on different datasets including multiple data modalities: images, videos, and single-view 3Ds. All the tasks share the same encoder. Here in order to learn from the different datasets and effectively utilize the knowledge from object segmentation to enhance the main task, we use a similar training protocol, where the interaction between both datasets and the joint representation gained from the shared encoder are learned intrinsically, similar to  \parencite{girdhar2022omnivore}. Meanwhile, because the objects in the saliency dataset are common in the object segmentation dataset (PASCAL VOC dataset covers most of the common objects in the real world and is the most famous benchmark in semantic segmentation), the knowledge from the segmentation decoder can be transferred effectively into saliency decoder. We show in section 4.5 that the segmentation decoder can also output the coarse result for a saliency input. The decoder of the semantic segmentation subnet is very simple. First, convolution computes the probability for each object class and then deconvolution recovers the size of the feature map. In addition, when training the decoder network of the semantic segmentation task, we change the gradients to ten times smaller when the backpropagation reaches the encoder network because our main task is saliency prediction and the encoder should learn more in the feature space of that field.

The two tasks are trained at the same time. The input of saliency prediction does not go through the whole subnet of semantic segmentation, and vice versa. Both tasks share the same Transformer encoder. We take advantage of the Pascal VOC dataset to learn the feature for scene perception and transfer it to the main task thanks to  categories of objects in Pascal VOC also covering plenty of objects in the saliency dataset. In Figure \ref{fig:archi}, the blue and red boxes represent the decoder nets of saliency prediction and semantic segmentation, respectively. In one iteration of the training stage, the losses for both tasks are computed for the two decoder nets respectively and simultaneously. The losses are added as a joint one for gradient backpropagation, similar to \parencite{girdhar2022omnivore}. In our experiments, the batch sizes for saliency prediction and semantic segmentation are set to 8 and 1, respectively. The interaction between the two tasks is dealt with by the MAM module. As for the inference stage, the input only contains the image from the saliency dataset and the output is the corresponding saliency map gained from the saliency decoder net. Our extensive comparison experiments in section 4.5 demonstrate the effectiveness of the proposed joint learning.
\subsection{Loss Function}
Our model mainly contains two tasks, including saliency prediction and semantic segmentation, so we need two specific loss functions respectively. Traditionally, we utilize MSE \parencite{christoffersen2004importance} Loss for saliency prediction, because each pixel of the output has a numeric attention score while we utilize Cross Entropy Loss \parencite{ruby2020binary} for semantic segmentation, which could be seen as a normal classification problem. The total loss function is formulated as the following:
\begin{equation}
    MSE = \frac{1}{n}\sum_{i}(\hat{y}-y)^2
\end{equation}
\begin{equation}
    CE = -\frac{1}{n}\sum_{i}\sum_{c=1}^ky^{ic}*log(p^{ic})
\end{equation}
\begin{equation}
    L = \sum_{i=1}^4MSE_{i}*\frac{1}{2^{i-1}} + \lambda*CE
\end{equation}
\emph{MSE}, \emph{CE}, and \emph{L} refer to MSE Loss, Cross Entropy Loss, and the total loss respectively. \emph{n} stands for the number of pixels in the image. \emph{\^{y}} and \emph{y} refer to the predicted value and the ground truth value. From the result saliency map of stage 1 to stage 4, we multiply each MSE loss with a factor decreasing by 0.5, which means the deeper stage contributes less to the learning objective. In Cross Entropy Loss, \emph{k} is the total number of object classes and \emph{y$^{ic}$} is 1 only if \emph{c} is the corresponding object class. \emph{$\lambda$} is the hyperparameter that controls the relative importance when training the multiple tasks. We set $\lambda$ to 0.1 in our experiments.
\section{Summary}
The main structure of our method is built upon multi-task learning across multiple datasets. The specific decoders for the tasks share the same Transformer encoder and interact through the Multi-task Attention Module. The loss functions of the tasks are added as a joint one, similar to \parencite{girdhar2022omnivore}. Our experiments in the following chapter verify the effectiveness of our ideas.

\chapter{Experiments}
\addtocontents{toc}{\protect\enlargethispage{\baselineskip}}
In this chapter, we would mainly discuss and demonstrate the effectiveness of our idea. We would do further and deeper analysis for our model. In addition, both quantitative and qualitative comparisons between our model and other methods would be presented. The code would be available at https://github.com/zsbluesky/Mcomp-at-NUS.
\section{Experimental Setup}
\subsection{Datasets and Evaluation Metrics}
To evaluate our model, we choose two benchmark datasets generally used in this field, SALICON \parencite{jiang2015salicon} and MIT1003, \parencite{Judd_2012} static image saliency prediction datasets. The SALICON dataset is the largest dataset in saliency prediction, containing 10,000 images for training and 5,000 images for validation and testing respectively. The human eye attention annotations are collected on Amazon Mechanical Turk through a general-purpose mouse. Mouse clicking is used to simulate the human gaze and participants to build the dataset click where they look in the images. Due to its large amount of data and diversity of objects and scenes in the images, SALICON has been widely used as the pre-training dataset before transferring onto other much smaller datasets. The training and validation upon SALICON are very stable. The MIT1003 dataset is a well-known benchmark dataset in the field of saliency prediction, containing only 1,003 images for training and 300 images (MIT300 test set) for online testing. The annotations are collected by the eye-tracking device. The result of the testing of both SALICON \parencite{jiang2015salicon} and MIT1003 datasets \parencite{Judd_2012} could be found on their official website. During training, we also take advantage of the PASCAL VOC 2012 dataset for the semantic segmentation branch. We use five metrics to evaluate our model effectively: AUC-Judd (AUC-J), Similarity Metric (SIM), shuffled AUC (s-AUC), Linear Correlation Coefficient (CC), Normalized Scanpath Saliency (NSS), Information Gain (IG) and KL Divergence (KL) \parencite{bylinskii2018different}. AUC-Judd and shuffled AUC measure a 2 alternative forced choice task, where the model should determine which one of two locations has been fixated. The non-fixation distribution is uniform for AUC-Judd and image independent center bias for shuffled AUC. NSS refers to the average saliency value of fixated pixels after the saliency map is normalized. IG compares the average log-probability of fixated pixels to that given by a center bias. Correlation Coefficient, KL Divergence, and Similarity Metric measure the closeness between the predicted saliency map and ground truth saliency map. The details of the metrics are cited from \parencite{bylinskii2018different}.

\subsection{Implementation Details}
First, we do training upon the largest saliency prediction dataset, SALICON, as in previous research. We use SGD \parencite{amari1993backpropagation} optimizer to train the network, with the momentum of 0.9 and the weight decay of 0.0001. The learning rate is set to 0.01. The batch size of the saliency prediction task is set to 8, while that of the semantic segmentation task is set to 1. The input image sizes of SALICON are the same, 640$\times$480, while that of PASCAL VOC vary. Mean Squared Loss (MSE loss) and Cross Entropy Loss are used as the training objectives for saliency prediction and semantic segmentation, respectively. \emph{$\lambda$} in Equation 4.13 is set to 0.1. Then we transfer the model onto the MIT1003 dataset. We use Adam \parencite{kingma2014adam} optimizer because of the smaller amount of data and the learning rate is set to 0.00001. Here we set $\lambda$ to 0 in the total loss function since the semantic segmentation branch has already been trained. The image sizes of MIT1003 are not the same, so we first do padding and then resize the images to 640$\times$480. In all training processes, stage decay is used for better learning and less overfitting. The number of Transformer encoders we use is 12 in order to make use of the pretrained Transformer from Imagenet \parencite{deng2009imagenet}. We start the training from the pretrained model from ImageNet (Resnet50+ViT) when compared with the state-of-the-art methods. The experiments are implemented on 2$\times$NVIDIA RTX A5000 GPUs (24564 MiB each).

\section{Quantitative Evaluation}
The quantitative evaluation of SALICON datasets is shown in Table \ref{tab:sali}. Here we include the recent state-of-the-art methods, which are also compared by other researchers: CEDNS \parencite{qi2019convolutional}, DI-Net \parencite{yang2019dilated}, EML-Net \parencite{jia2020eml}, SAM-ResNet \parencite{cornia2018predicting}, GazeGAN \parencite{che2019gaze} and UNISAL \parencite{droste2020unified}. All the results are from the online system\footnote{http://salicon.net/challenge-2017/}  since we do not have the test dataset. From Table \ref{tab:sali}, we could see that our model performs best, with the best scores in four metrics: s-AUC, AUC-Judd, IG, and KL. EML-Net is the highest in NSS while DI-Net is the highest in CC and SIM. The performance of our model in AUC metrics means that our method is stronger in 2AFC (2 alternative forced choice) and yields the highest performance in the result saliency map \emph{$p_{fix}(x, y)/p_{nonfix}(x, y)$}, where \emph{$p_{fix}(x, y)$} refers to the probability of fixation at the point and \emph{$p_{nonfix}(x, y)$} refers to the probability of nonfixation at the point. The performance of our model in KL metric shows that the distribution of our result saliency map is more similar to the ground truth.

The quantitative evaluation of the MIT300 dataset is shown in Table \ref{tab:mit}, where the IG metric is not provided by the benchmark. Red stands for the first place while blue stands for the second place. We compare our model with the recent methods: DVA \parencite{wang2017deep}, SalGAN \parencite{pan2017salgan}, SAM-Resnet \parencite{cornia2018predicting}, CASNet \cite{fan2018emotional}, GazeGAN \parencite{che2019gaze}, AttentionInsight\footnote{https://attentioninsight.com/.},  TransalNet \parencite{lou2021transalnet} and EML-Net \parencite{jia2020eml}.  Here we do not include the probabilistic models as in \parencite{lou2021transalnet}, since the result saliency maps from probabilistic models would be changed to metric-specific results, which would result in the traditional saliency models performing suboptimally. From Table \ref{tab:mit}, our method performs competitively. Our score of AUC-Judd is very close to the highest and our method far outperforms all other models in the s-AUC metric.

\begin{table}[h!]

\caption{Comparison with the state-of-the-art methods on SALICON dataset.}
\centering
\begin{tabular}{ |p{2.5cm}||p{1.5cm}|p{1.5cm}|p{1.0cm}|p{1.0cm}|p{1.0cm}|p{1.0cm}|p{1.0cm}|}
 \hline
 \multicolumn{8}{|c|}{SALICON Saliency Prediction Challenge (LSUN 2017)} \\
 \hline
 Method &s-AUC & AUC-Judd &IG&NSS&CC&SIM&KL\\ 
 \hline
 CEDNS	&	0.745	&0.862&	0.357&	2.045&	0.862&	0.753&	1.026\\
 DI-Net	&	0.739&	0.862&	0.195&	1.959&	\textbf{0.902}&	\textbf{0.795}&	0.864\\
 EML-Net &		0.746&	0.866&	0.716&	\textbf{2.050}&	0.866&	0.78&	0.52\\
 SAM-ResNet	&	0.741&	0.865&	0.538&	1.99&	0.899&	0.793&	0.61\\
 GazeGAN&		0.736&	0.864&	0.720&	1.899&	0.879&	0.773&	0.376\\
 UNISAL&		0.739&	0.864&	0.777&	1.952&	0.879&	0.775&	0.346\\ 
 \textbf{Ours}& \textbf{0.747}&\textbf{0.867}&\textbf{0.806}&1.968&0.880&0.749&\textbf{0.265}\\  
 \hline
\end{tabular}
\label{tab:sali}
\end{table}

\begin{table}[!htbp]
\centering
\caption{Comparison with the state-of-the-art methods on MIT300 dataset.}
\begin{tabular}{ |p{3.0cm}||p{1.5cm}|p{1.5cm}|p{1.0cm}|p{1.0cm}|p{1.0cm}|p{1.0cm}|}
 \hline
 \multicolumn{7}{|c|}{MIT300 Dataset} \\
 \hline
 Method &s-AUC & AUC-Judd &NSS&CC&KL&SIM\\
 \hline
 DVA&	0.7257&0.8430&	1.9305&	0.6631&	\textcolor[rgb]{0,0,1}{0.6293}&	0.5848\\
 SalGAN&	0.7354&0.8498&	1.8620&	0.6740&	0.7574&	0.5932\\
 SAM-Resnet&	0.7396&0.8526&	2.0628&	0.6897&	1.1710&	0.6122\\
 CASNet &	0.7398&0.8552&	1.9859&	0.7054&	\textcolor[rgb]{1,0,0}{0.5857}&	0.5806\\
 GazeGAN&	0.7316&0.8607&	2.2118&	0.7579&	1.3390&	0.6491\\
 AttentionInsight &	0.7446&0.8640&	2.1825&	0.7578&	0.8873&	0.6551\\
 TranSalNet &	\textcolor[rgb]{0,0,1}{0.7471}&0.8730&	2.3758&	\textcolor[rgb]{1,0,0}{0.7991}&	0.9019&	\textcolor[rgb]{1,0,0}{0.6852}\\
 EML-Net &	0.7469&		\textcolor[rgb]{1,0,0}{0.8762}&	\textcolor[rgb]{1,0,0}{2.4876}&	0.7893&0.8439&\textcolor[rgb]{0,0,1}{0.6756}\\
 \textbf{Ours}&	\textcolor[rgb]{1,0,0}{0.7549}&\textcolor[rgb]{0,0,1}{0.8744}&	\textcolor[rgb]{0,0,1}{2.3762}&	\textcolor[rgb]{0,0,1}{0.7897}&	0.7146&	0.5313\\
 \hline
\end{tabular}
\label{tab:mit}
\end{table}

\section{Qualitative Evaluation}
\begin{figure}
    \centering
    \centerline{\includegraphics[scale=0.5]{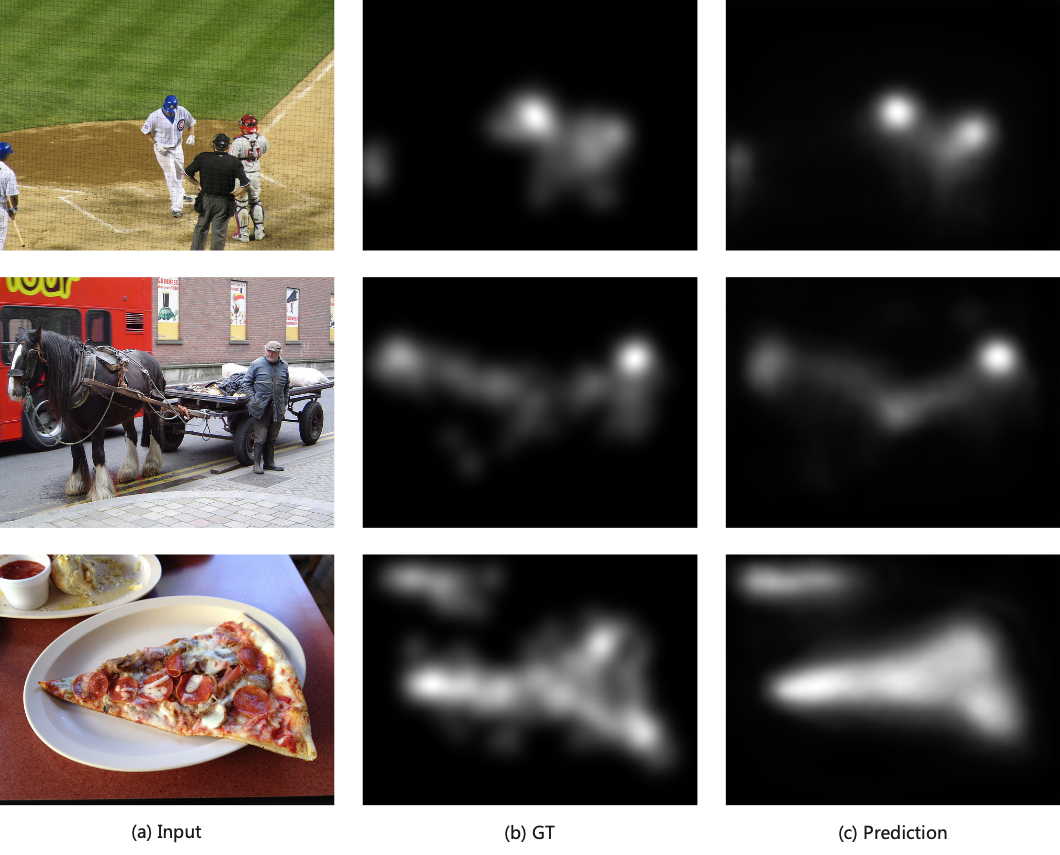}}
    \caption{Sample qualitative results on SALICON.}
    \label{fig:quasal1}
\end{figure}

\begin{figure}
    \centering
    \centerline{\includegraphics[scale=0.5]{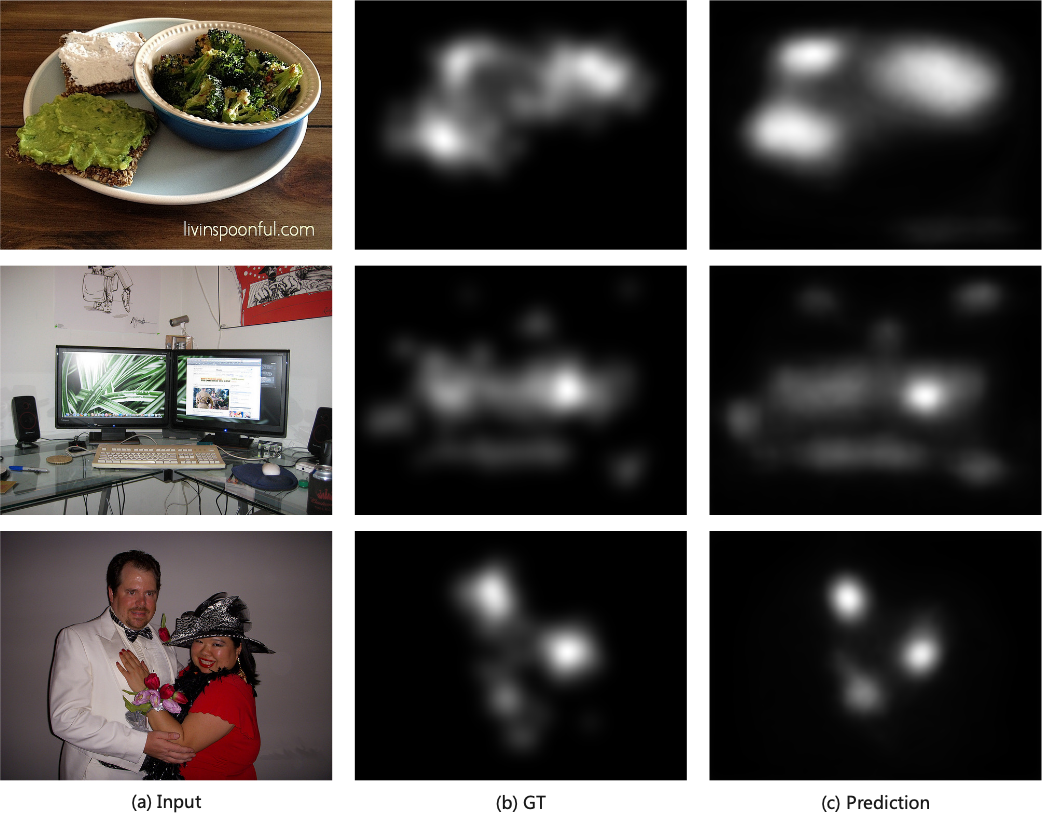}}
    \caption{Sample qualitative results on SALICON.}
    \label{fig:quasal2}
\end{figure}

\begin{figure}
    \centering
    \centerline{\includegraphics[scale=0.6]{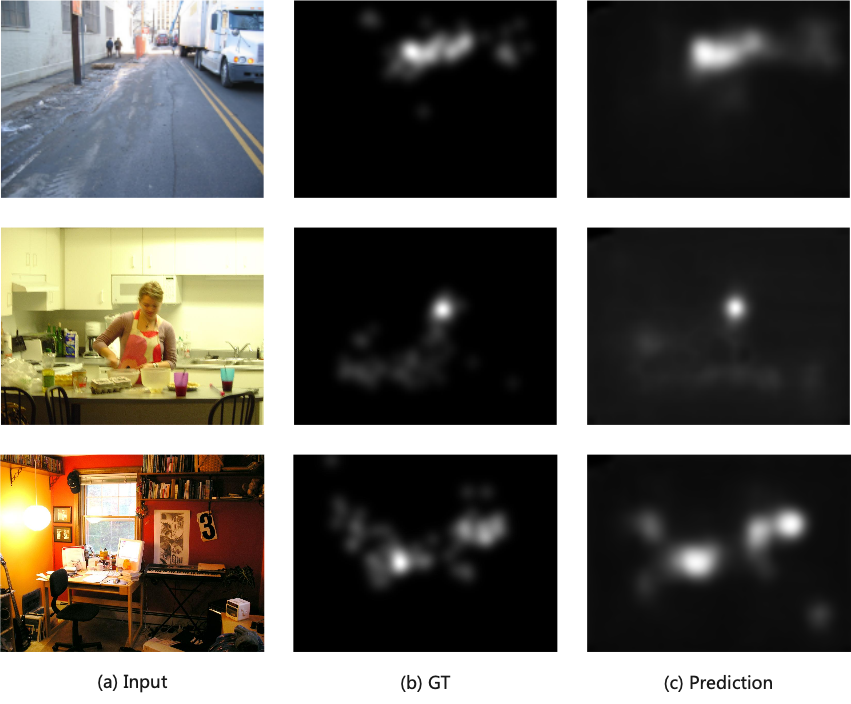}}
    \caption{Sample qualitative results on MIT1003.}
    \label{fig:quamit1}
\end{figure}

\begin{figure}
    \centering
    \centerline{\includegraphics[scale=0.45]{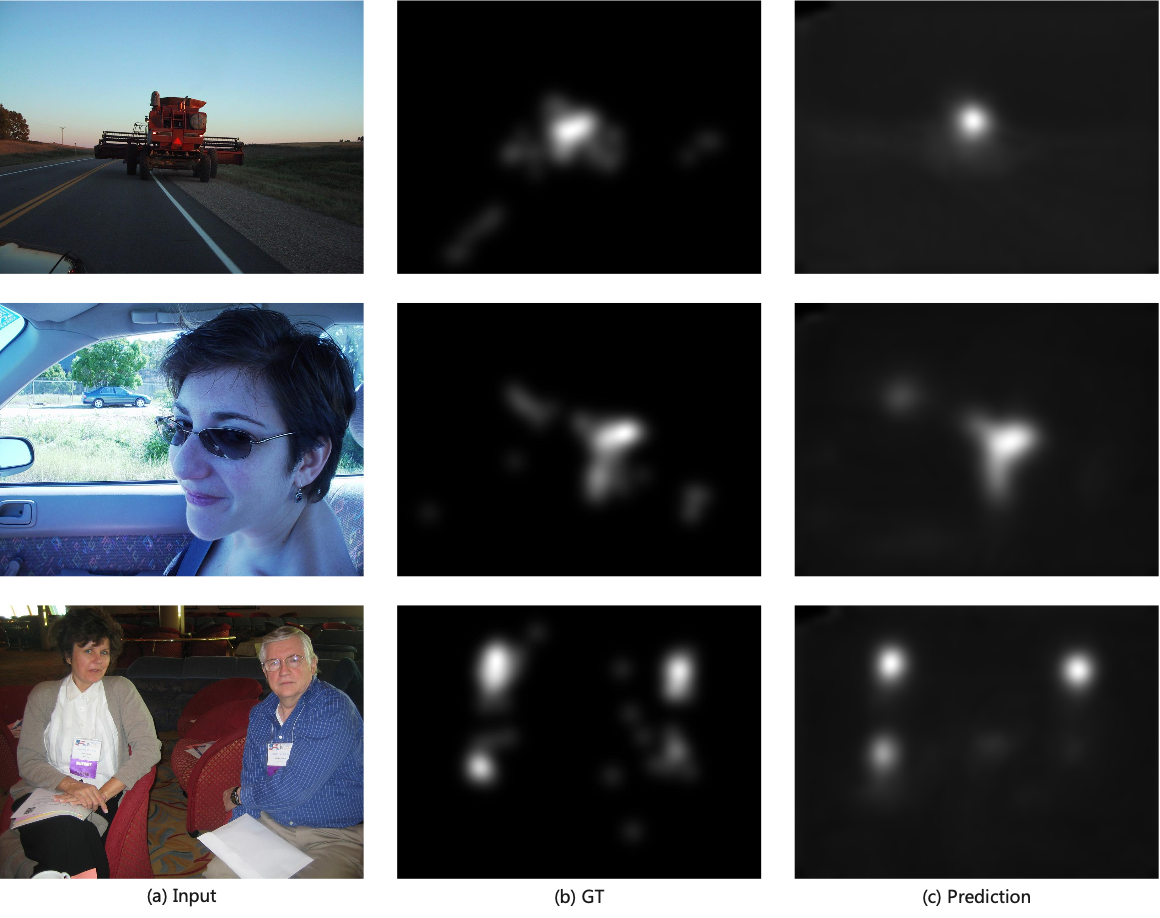}}
    \caption{Sample qualitative results on MIT1003.}
    \label{fig:quamit2}
\end{figure}

The qualitative results of SALICON are shown in Figure \ref{fig:quasal1} and Figure \ref{fig:quasal2}. Our model could predict well the human attention in both common and complex scenes. The salient regions could be recognized precisely. We could find that when there are people or faces in the image, the saliency could be more obvious. 
The result saliency maps on MIT1003 are shown in Figure  \ref{fig:quamit1} and Figure \ref{fig:quamit2}. The salient regions seem more clear than the SALICON dataset. Our model performs well compared with the ground truth.
\section{Ablation Study}
\begin{figure}
    \centering
    \centerline{\includegraphics[scale=1.0]{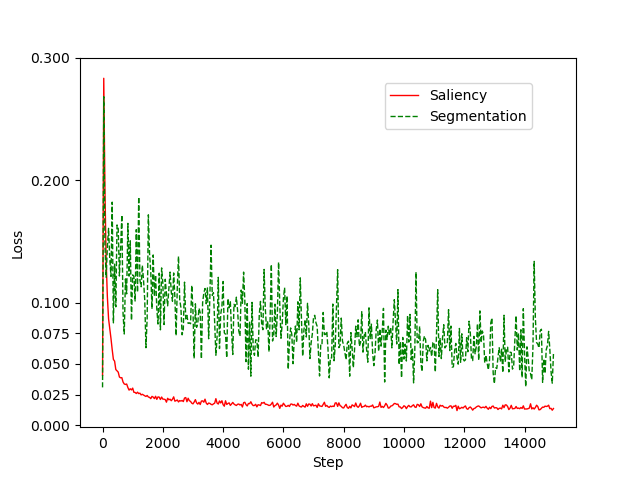}}
    \caption{Losses for the final multi-task training. The green and red lines represent segmentation loss and saliency prediction loss, respectively. The segmentation loss is multiplied by 0.1, which is the weight demonstrated in our proposed method, for better visualization. Both losses decrease smoothly.}
    \label{fig:loss}
\end{figure}

In this section, we would demonstrate the effectiveness of the modules we propose. The result of our ablation experiments is shown in Table \ref{tab:ablation}. Red refers to the first place while blue refers to the second. The baseline model is based on pretrained Resnet50 from ImageNet. We concatenate the features from all stages and pass through convolution to get the saliency result. The decoder in Table \ref{tab:ablation} refers to the subnet in 3.4 without global feature concatenation (skip connection). The saliency decoder network contributes to a considerable performance increase. We could explain the effectiveness of skip connection that the model learns more from the global cues than the traditional feature combination. The Transformer module yields the largest gain thanks to its strong image representation and global perception by multi-head self-attention. From Table \ref{tab:ablation}, simply adding another task without any interaction between the multiple tasks could lead to a performance decrease. The model may not focus on the main task during the training process and another task could be an interference. So we design the Multi-task Attention Module (MAM) to introduce the feature from the segmentation branch as feature selection for our saliency network, which improves the performance to a large extent. Another interesting thing we find is that multi-task learning increases the scores on both CC and SIM, which are the metrics measuring whether the model is close to human perception \parencite{lou2021transalnet}. This further demonstrates object segmentation could assist the saliency prediction. In addition, multi-supervision learning could be useful because of more objectives to enhance the learning. The training process of  multi-task learning is shown in Figure \ref{fig:loss}.

\begin{table}[!htbp]
\centering
\caption{Ablation study of our proposed method upon the SALICON validation set. The proposed components are added incrementally to the baseline model to quantify their contributions.}
\begin{tabular}{ |p{3.5cm}||p{1.5cm}|p{1.5cm}|p{1.0cm}|p{1.0cm}|p{1.0cm}|}
 \hline
 \multicolumn{6}{|c|}{SALICON validation set} \\
 \hline
 Config. &AUC-Judd&s-AUC&NSS&CC&SIM\\
 \hline
 Resnet50&	0.759&	0.646&1.214&0.824&\textcolor[rgb]{1,0,0}{0.740}\\
 + Decoder 	&0.761	&0.649&1.219&0.825&0.718\\ 
 + Skip Connection&0.764&0.650&1.230&{0.836}&0.722\\
 + Multi-supervision&0.765&0.651&1.244&\textcolor[rgb]{0,0,1}{0.840}&0.723\\
 + Transformer	&\textcolor[rgb]{0,0,1}{0.772}	&\textcolor[rgb]{0,0,1}{0.655}&\textcolor[rgb]{0,0,1}{1.295}&0.824&0.695\\
 + Multi-task& 0.771&0.654&1.293&\textcolor[rgb]{1,0,0}{0.858}&\textcolor[rgb]{0,0,1}{0.738}\\
 + MAM& \textcolor[rgb]{1,0,0}{0.774}&\textcolor[rgb]{1,0,0}{0.656}&\textcolor[rgb]{1,0,0}{1.302}&\textcolor[rgb]{0,0,1}{0.840}&0.736\\
 
 \textbf{Final}& \textcolor[rgb]{1,0,0}{0.774}&\textcolor[rgb]{1,0,0}{0.656}&\textcolor[rgb]{1,0,0}{1.302}&\textcolor[rgb]{0,0,1}{0.840}&0.736\\
 \hline
\end{tabular}
\label{tab:ablation}
\end{table}

\section{Consistency between saliency and object perception}
\begin{figure}
    \centering
    \centerline{\includegraphics[scale=0.3]{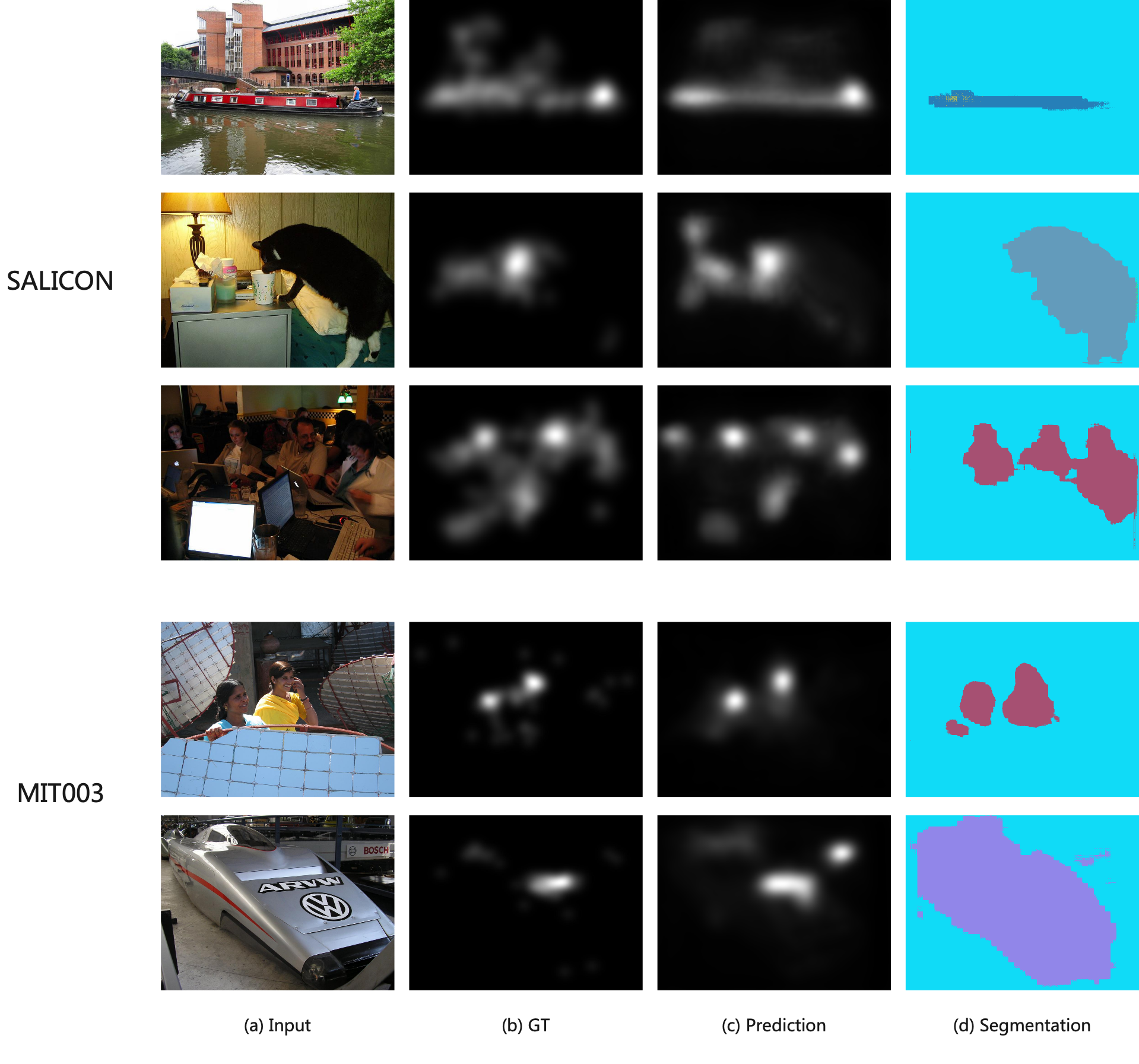}}
    \caption{Consistency between saliency and object perception. Human beings tend to focus on the main common objects in a scene. The eye fixations are closely related to object segmentation. (c) and (d) stand for our saliency predictions and object segmentations. The results of the MIT1003 dataset are from the model trained from scratch.}
    \label{fig:consis}
\end{figure}

Human beings tend to focus on the main common objects in a scene, which is shown in Figure \ref{fig:consis}. The test results are from the SALICON and MIT1003 datasets. The segmentation results are a little coarse since we do not employ a complex structure. Obtaining good performances in both tasks could be future work. Eye fixations are closely related to object perception. In a deep learning model, one way to build this relation is by multi-task learning. To further justify our idea that object perception could be a guidance for human saliency prediction, we train our network from scratch upon the MIT1003 dataset. We randomly split the dataset into the training set with 903 images and validation set with 100 images. The comparison experiment result is shown in Table \ref{table:consis}. The joint learning of segmentation greatly benefits the saliency prediction from the quantitative metrics, especially in CC and SIM metrics, which measure the closeness between the model and human eyes \parencite{lou2021transalnet}. From Table \ref{table:consis}, the performance drops if the semantic segmentation branch is eliminated. Furthermore, if the segmentation branch remains but $\lambda$ equals to 0, the performance is even worse, which strongly and forcefully demonstrates that only attaching the MAM module onto the saliency network is not effective. We could conclude that the effectiveness of the MAM module comes from the consistency between saliency prediction and object perception. In addition, as shown in Table \ref{table:consis2}, we verify the positive effect of our multi-task learning in transfer learning. Our model is first pretrained on SALICON and then finetuned upon MIT1003. The training and validation sets of the MIT1003 dataset are the same as in Table \ref{table:consis}. The same methods for both datasets in Table \ref{table:consis2} mean each pair shares the same model architecture during pretraining and finetuning. We could find that in the SALICON dataset, semantic segmentation learning improves the model performance in four metrics, and also if $\lambda$ is set to zero, the score of s-AUC drops dramatically, which is the main metric in SALICON Challenge 2017. During the finetuning stage, because the size of the dataset is much smaller, the result scores shake to a larger extent. From the validation result, the model gains a huge leap in s-AUC, which shows the effectiveness of the object perception.
\begin{table}[h!]
\centering
\caption{Comparison experiments upon MIT1003 to further demonstrate the idea that object perception could be  guidance for saliency prediction. The results are from the model trained from scratch.}
\begin{tabular}{c c c c c c} 
 \hline
 Method & AUC-Judd & s-AUC & NSS&CC&SIM \\ 
 \hline\hline
 w/o seg. branch&0.856&0.742&1.823&0.493&0.415\\
 w/ seg. branch, $\lambda$ = 0&0.845&0.730&1.660&0.454&0.341\\
 \textbf{ours} &\textbf{0.869}&\textbf{0.748}&\textbf{1.939}&\textbf{0.528}&\textbf{0.426}\\

 \hline
\end{tabular}
\label{table:consis}
\end{table}

\begin{table}[h!]
\centering
\caption{Comparison experiments to justify the effect of multi-task learning in transfer learning. The model is first pretrained on SALICON and transferred onto the MIT1003 dataset. The same methods in both datasets mean each pair uses the same model architecture in pretraining and finetuning.}
\begin{tabular}{c c c c c c c} 
 \hline
 Dataset & Method & AUC-J & s-AUC & NSS&CC&SIM \\ 
 \hline\hline
 &w/o seg. branch&0.772&0.655&1.295&0.824&0.695\\
 SALICON&w/ seg. branch, $\lambda$=0&0.773&0.652&1.280&\textbf{0.865}&0.734\\
 &\textbf{ours} &\textbf{0.774}&\textbf{0.656}&\textbf{1.302}&0.840&\textbf{0.736}\\
 \hline
 &w/o seg. branch&0.911&0.804&\textbf{2.890}&0.752&0.350\\
 MIT1003&w/ seg. branch, $\lambda$=0&0.908&0.809&2.864&0.749&0.364\\
 (finetuning)&\textbf{ours} &\textbf{0.913}&\textbf{0.815}&2.881&\textbf{0.755}&\textbf{0.365}\\
 \hline
\end{tabular}
\label{table:consis2}
\end{table}

\section{Feature Map Analysis}
\begin{figure}
    \centering
    \centerline{\includegraphics[scale=0.5]{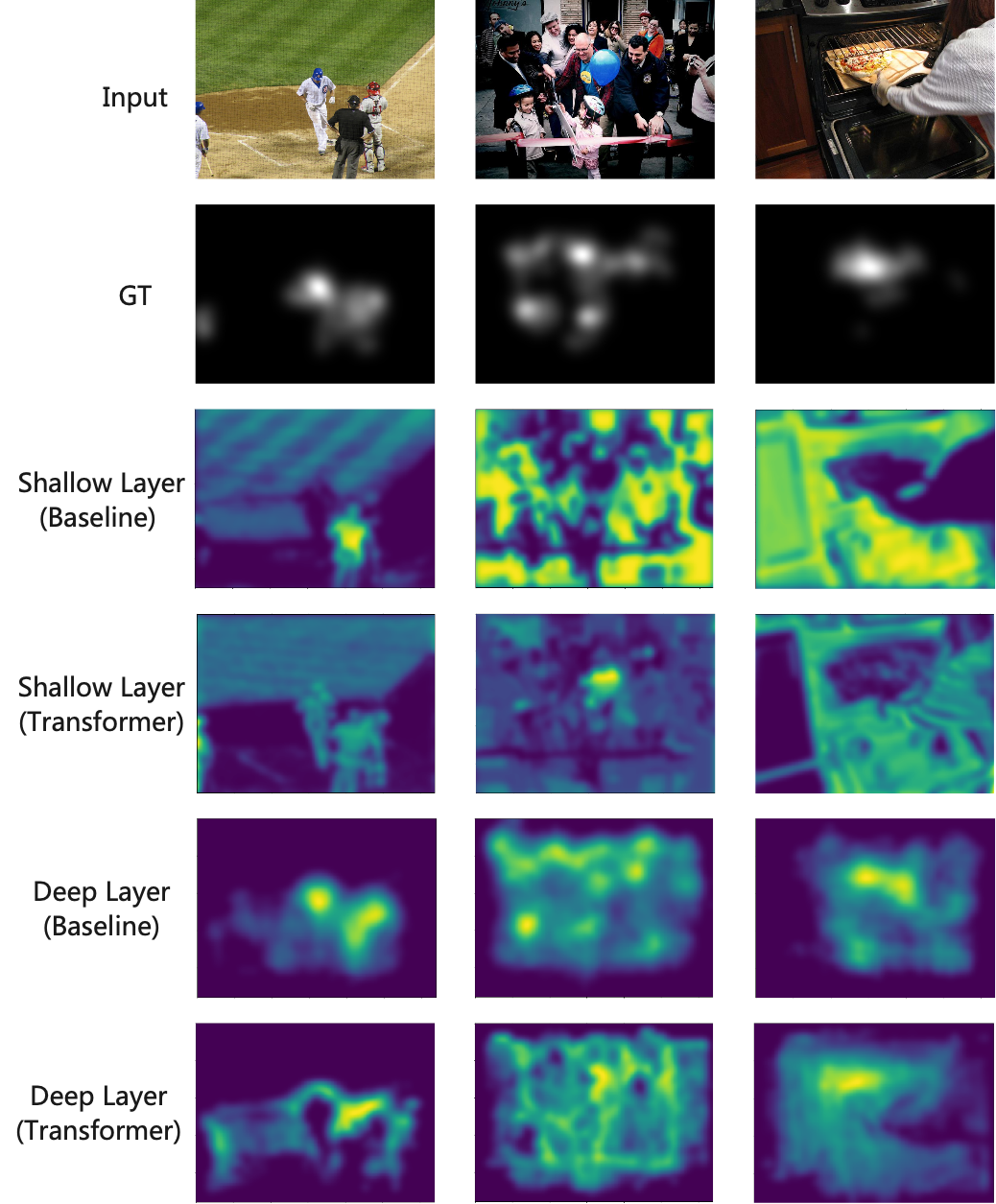}}
    \caption{Feature map analysis from the encoder. The baseline model is the basic Resnet. The Transformer helps to recognize more global cues and locate the salient regions more precisely.}
    \label{fig:former}
\end{figure}

In this section, we would analyze the feature maps from the encoder network. The visualization is shown in Figure \ref{fig:former}. The baseline model is the basic Resnet. We could find that the baseline model tends to learn the features from specific objects, e.g., people or faces, and could not locate the salient regions precisely if it comes to a complex scene. Also, due to fewer global cues, the baseline model may lose other detail in the input image. The Transformer-based encoder could enhance the global information of the image, which could be demonstrated in Figure \ref{fig:former} that more objects and more background information are attended. In addition, the Transformer encoder could better recognize the salient regions in a complex scene through either shallow layers or deep layers. We should also consider the combination of the shallow features and deep features since it is implemented in the decoder network. It is easy to be found that the addition of shallow features and deep features from the baseline model is further from the ground truth, leading to less effective learning for saliency prediction.

\section{Failure Case Discussion}
\begin{figure}
    \centering
    \centerline{\includegraphics[scale=0.8]{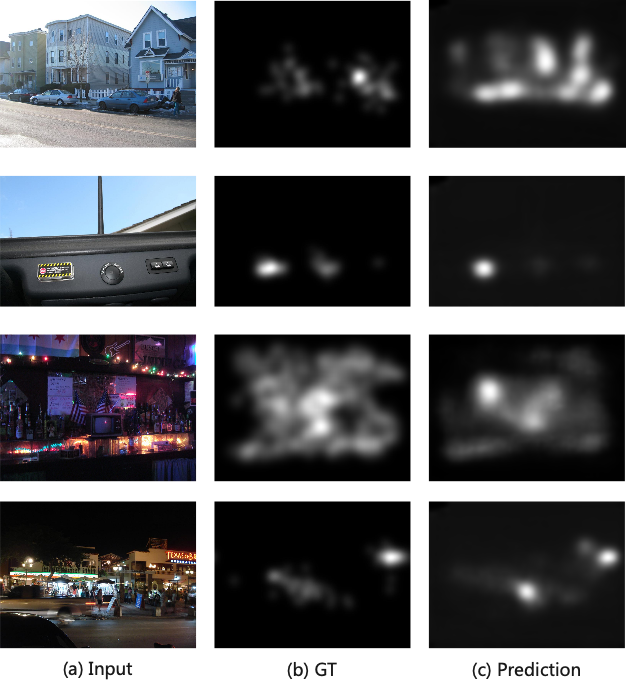}}
    \caption{Failure case discussion.}
    \label{fig:badcase}
\end{figure}

Although our model and many other state-of-the-art methods achieve very high scores in AUC-Judd or s-AUC metrics on multiple benchmarks, there are still difficult cases where the result saliency maps have errors that are not incidental. Some failure cases in Figure \ref{fig:badcase} could be summarized as: a) the model fails to recognize the salient regions when they are not very obvious and the result saliency is assigned to almost the whole picture, b) the model only focuses on the most salient objects but ignore others, caused by the overfitting of the training data, c) the salient regions could not be located in a very complex or dark scene. As for the scenes which are very complex or have no obviously salient region, one way we propose to solve the problem is by constructing a larger saliency dataset which is much closer to reality. Few datasets satisfy the three conditions: a) large-scale, b) built with an eye-tracking device, and c) including lots of complex scenes which are closer to our surroundings. In the field of Salient Object Detection (SOD), some research finds that the state-of-the-art methods achieving high scores on traditional benchmarks perform very badly when they are deployed on mobiles \parencite{fan2018salient}, so SOC (Salient Object in Clutter) dataset \parencite{fan2018salient} is proposed to solve the problem. A new saliency prediction benchmark closer to reality is also needed. On the other hand, we could not enlarge the deep learning model when the amount of data is limited. As for the problem that the model could not recognize the less salient regions, we could enhance the image representation by model ensembling \parencite{linardos2021deepgaze} or using more Transformer blocks.
\section{Summary}
In this chapter, we introduce the benchmark datasets and the implementation of our method. The quantitative and qualitative results demonstrate our model achieves competitive performance against the recent state-of-the-art methods. The ablation study demonstrates the effectiveness of the modules we proposed. In addition, we visualize the feature maps from the Transformer and the traditional Resnet, which show that the Transformer could capture more global cues. The consistency between object segmentation and saliency prediction is verified both by finetuning and training from scratch. At last, we show some common failure cases in the saliency prediction of our method and propose some possible solutions for them.

\chapter{Conclusion}
\addtocontents{toc}{\protect\enlargethispage{\baselineskip}}
\section{Summary}
In this thesis, we design a Transformer-based network which includes unified learning of semantic segmentation to predict human attention. Transformer could introduce global features, which benefits the predicting process since people tend to get a global view of the scene when determining which areas are salient in their brains. The main structure of our network makes use of classical bottom-up multi-level feature fusion, combining the global cues from the  Transformer and local cues from traditional CNN. Inspired by human perception when looking at a picture, we introduce another subnet for semantic segmentation. This simulates the human perception process where recognizing objects in the picture 
contributes to distributing attention to it. Our experiments show the effectiveness of 
the proposed ideas. Our model achieves competitive performance compared with other state-of-the-art methods.
\section{Future Work}
Our model simply employs the Transformer encoder to get the global cue from the deepest feature of the traditional CNN, so a task-specific Transformer module could be proposed to further enhance the performance. As for the multi-task learning, our subnet of semantic segmentation is simple and the qualitative results are still not perfect. A comprehensive model could be proposed, which could achieve competitive performances in multiple tasks. Also, other closely related tasks for scene understanding may be useful, e.g., depth estimation, object detection, etc.

Through the analysis of our result saliency maps, we find that there are still some failure cases which are not incidental. When it comes to a more complex scene, our model may fail. Few datasets built with eye tracking devices contain enough complex scenes closer to reality, so a new large-scale benchmark could be proposed. The MIT1003 dataset only contains 1, 003 images and this is not adequate if we continue to enlarge our model.

\newpage

\printbibliography


\appendix
\chapter{Quantitative and qualitative results on CAT2000}
In this appendix, we would show our quantitative result compared with other state-of-the-art methods on CAT2000 \parencite{borji2015cat2000} and our qualitative results for evaluation. Our model achieves a very competitive performance and improves the state-of-the-art on many metrics. CAT2000 dataset is a free viewing eye fixation dataset,  containing 4,000 images (2,000 for training and 2,000 for online testing, respectively) from 20 different categories, e.g. satellite, outdoor, social, etc. During the offline evaluation, we randomly select 200 images from the whole training set. The size of the input image is very large, 1920x1080, and we resize it to 640x480 during training and testing. The implementation is the  same as MIT1003 where we finetune on the CAT2000 dataset with the pretrained model from SALICON.
\section{Quantitative comparison with the state-of-the-art methods}
Table \ref{tab:cat} shows the quantitative result of the CAT2000 dataset. The result is from the benchmark website \parencite{Judd_2012}. Here we also do not include the probabilistic models whose metric calculation is optimized. We compare our method with DVA \parencite{wang2017deep}, SalGAN \parencite{pan2017salgan}, FES \parencite{rezazadegan2011fast}, LDS \parencite{fang2016learning}, CovSal \parencite{erdem2013visual} and  eDN \parencite{vig2014large}. Our method significantly improves the state-of-the-art in almost all the metrics. 
\begin{table}[!htbp]
\centering
\caption{Comparison with the state-of-the-art methods on the CAT2000 dataset.}
\begin{tabular}{ |p{3.0cm}||p{1.5cm}|p{1.5cm}|p{1.5cm}|p{1.5cm}|p{1.5cm}|p{1.5cm}|}
 \hline
 \multicolumn{7}{|c|}{CAT2000 Dataset} \\
 \hline
 Method &AUC-Judd & s-AUC &NSS&CC&KL&SIM\\
 \hline
 DVA&0.8000&	0.6253&	1.4287&	0.5033&	{0.9931}&	0.4924\\
 SalGAN&0.8085&	0.6354&	1.4624&	0.5207&	1.1155&	0.4921\\
 FES&0.8212&	0.5450&	1.6103&	0.5799&	2.6123&	0.5255\\
 LDS &	0.8281&	0.5669&	1.5356&	{0.5574}&	1.0453&0.5267\\
 CovSal&0.8402&	0.5570&	1.7411&	0.6251&	1.5630&	\textbf{0.5603}\\
 eDN &0.8470&	0.5782&	1.2092&	0.4482&	1.1856&	0.3997\\
 \textbf{Ours}&\textbf{0.8771}&	\textbf{0.6005}&	\textbf{2.3868}&	\textbf{0.8408}&	\textbf{0.6978}&	0.5363\\
 \hline
\end{tabular}
\label{tab:cat}
\end{table}

\section{Qualitative Evaluation}
\begin{figure}
    \centering
    \centerline{\includegraphics[scale=0.5]{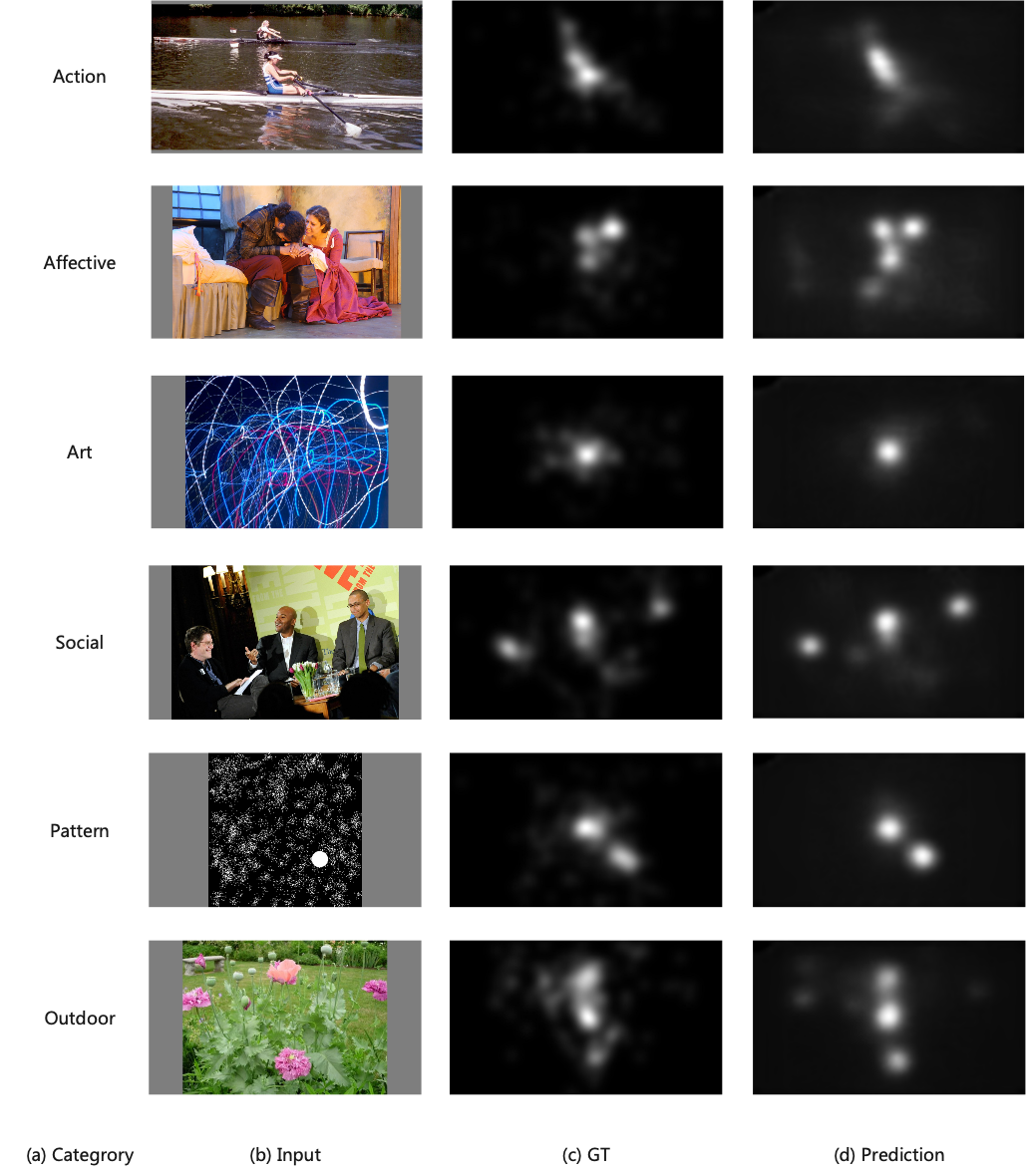}}
    \caption{Sample qualitative results on CAT2000.}
    \label{fig:cat}
\end{figure}

Here we provide the qualitative results on CAT2000, shown in Figure \ref{fig:cat}. The samples are chosen from our self-defined evaluation set, from six different categories of CAT2000. The most salient regions or objects could be captured but less important positions are not well predicted. The ground truth results are generally consistent with our predictions.
\chapter{More Demonstrations about Object Perception in Saliency Prediction}
In this section, we will give more demonstrations about object perception in saliency prediction.
\begin{figure}
    \centering
    \centerline{\includegraphics[scale=0.5]{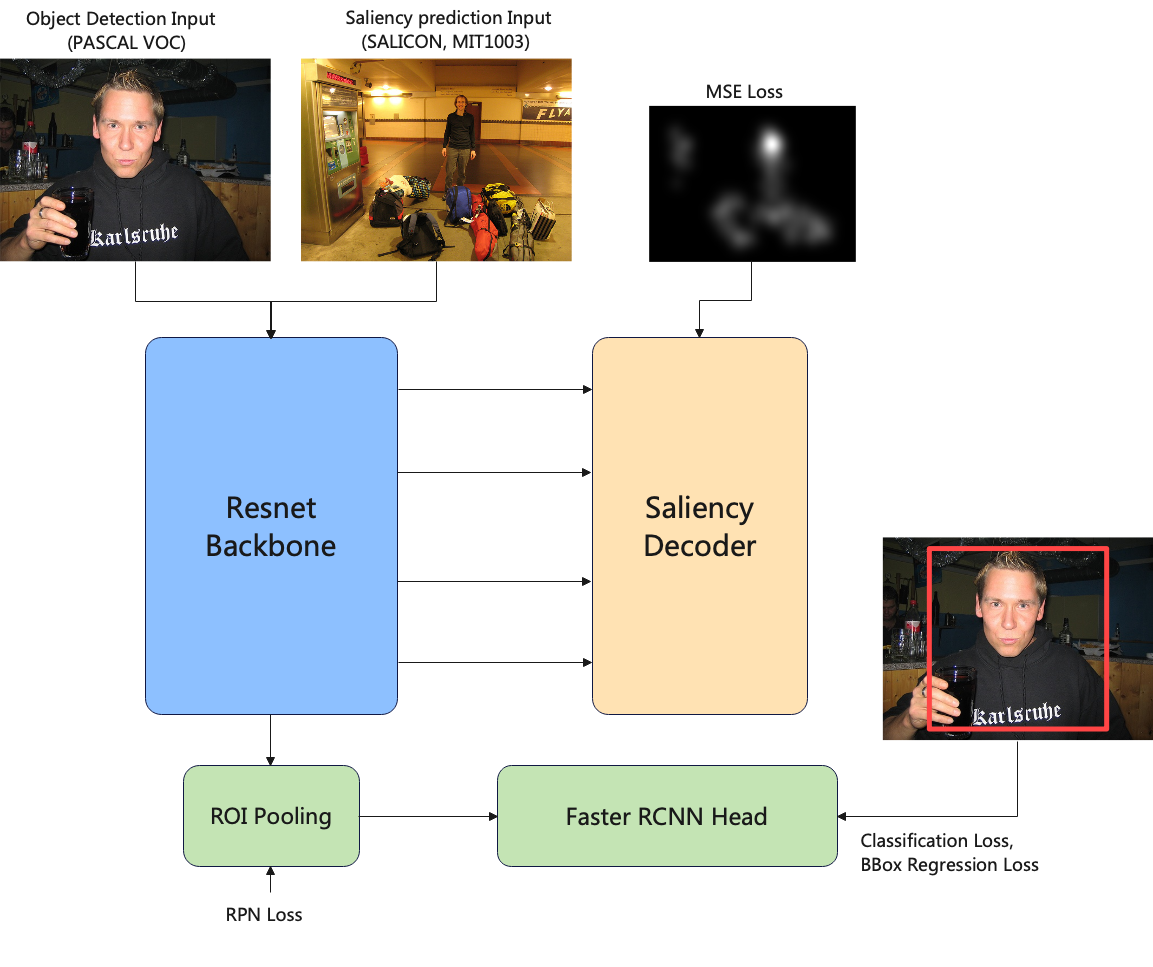}}
    \caption{Principle of the joint learning of saliency prediction and object detection. Here we compare the effects of different methods of object perception and build object detection subnet upon the baseline model in our ablation experiment. The saliency decoder is the same as the baseline model. Faster RCNN \parencite{ren2015faster} is utilized for object detection. Due to the ROI Pooling, we cannot gain the feature for the whole image so the proposed MAM module is not presented here.}
    \label{fig:detect}
\end{figure}
Specifically, we compare the effects of semantic segmentation and object detection on saliency prediction. Here we will only compare the effects of the joint learning, so we build the object detection and segmentation subnet upon the same baseline model in our ablation experiment. The principle of the joint learning of saliency prediction and object detection is shown in Figure \ref{fig:detect}. The decoder for saliency is the same as the baseline model in our ablation experiment, where the features of four stages from the encoder backbone are concatenated and go through one convolution to get the saliency result. Similar to our proposed architecture in Chapter 3, we utilize the Pascal VOC object detection dataset for joint learning. Faster RCNN \parencite{ren2015faster} is used for the learning of object detection. Both tasks share the same Resnet encoder. Due to the ROI Pooling, we cannot gain a representation for the whole image but separate objects, so the MAM module is not
\begin{table}[!htbp]
\centering
\caption{Comparison between different methods for object perception. Object perception is added separately. Object segmentation is built upon the same baseline model.}
\begin{tabular}{ |p{4.5cm}||p{1.5cm}|p{1.5cm}|p{1.5cm}|p{1.0cm}|p{1.0cm}|}
 \hline
 \multicolumn{6}{|c|}{SALICON Validation dataset} \\
 \hline
 Method &AUC-Judd & s-AUC &NSS&CC&SIM\\
 \hline
 Resnet50&0.759&	0.646&	1.214&	0.824&	0.740\\
 + Object Detection&0.756&0.645&1.208&\textbf{0.830}&0.748\\
 + Object Segmentation&\textbf{0.761}&\textbf{0.650}&\textbf{1.225}&0.826&\textbf{0.749}\\
 \hline
\end{tabular}
\label{tab:detect}
\end{table}
presented here. The quantitative result is shown in Table \ref{tab:detect}. We can find that the proposed object segmentation outperforms the object detection. The main reason is that due to the ROI Pooling in Faster RCNN, the features for the objects are separated in the computation of the detection head, hence the interaction between the multi-task learning is not well modeled due to the lack of the feature for the whole image. And the knowledge in Faster RCNN head cannot be well transferred because it only focuses on a single object (bounding box regression, classification). Our proposed method outperforms object detection in four metrics, especially in AUC-Judd and s-AUC.

\end{document}

%% file: ttlpg-kcl.tex

\thispagestyle{empty}

\noindent\begin{minipage}[c][\textheight]{\textwidth}
\centering
\singlespace

\large
\bfseries
\bigskip
\bigskip
\bigskip
\bigskip

SEMANTIC SEGMENTATION ENHANCED TRANSFORMER MODEL FOR HUMAN ATTENTION PREDICTION
\bigskip

\includegraphics[scale=0.6]{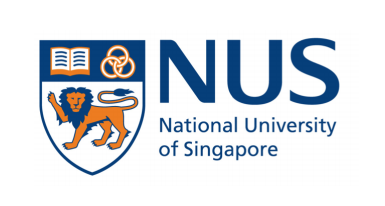}

SHUO ZHANG\\
\emph{(B.Eng., Nankai University)}
\mdseries


\vspace*{\fill}
\bfseries
A THESIS SUBMITTED

FOR THE DEGREE OF MASTER OF COMPUTING
in\\
COMPUTER SCIENCE\\
\bigskip
\bigskip
SCHOOL OF COMPUTING\\
NATIONAL UNIVERSITY OF SINGAPORE

\bigskip
\bigskip
2022

\vspace*{\fill}

\normalsize
\mdseries

\end{minipage}
\newpage

%% file: dclrtn.tex
\singlespace
\setcounter{page}{1}
\pagenumbering{roman}
\vspace*{\fill}

\begin{minipage}[c]{0.85 \textwidth}
\centering
\bfseries
DECLARATION
\bigskip

\normalsize
\large
\mdseries
I hereby declare that the thesis is my original work and it has been written by me in its entirety. I have duly acknowledged all the sources of information which have been used in the thesis.
\bigskip

This thesis has also not been submitted for any degree in any university previously.\par
\bigskip

\bigskip
\bigskip
\bigskip
\bigskip
\rule{1.5in}{0.75pt}

25 May 2022
\mdseries
\end{minipage}
\vspace*{\fill}
\newpage

%% file: acknwldgmnts.tex
\vspace*{\fill}
\noindent\begin{minipage}[c]{\textwidth}

\bfseries
\huge Acknowledgments
\bigskip

\normalsize
\mdseries
\noindent 
Here I appreciate my main supervisor Professor Mohan Kankanhalli and my co-supervisor Dr. Shaojing Fan for their guidance on the direction of my dissertation. I feel grateful to them when they give some excellent advice on my idea,  experiments, and anything else waiting to be explored in my study. I also appreciate other researchers in N-CRiPT who help me get familiar with the development environment and fix mechanical problems. \\

In addition, I appreciate my family and my friends who encourage me to overcome difficulties and complete the dissertation. 
\hspace{0.65 cm} 
\end{minipage}
\vspace*{\fill}
\newpage

%% file: smmry.tex
\vspace*{\fill}
\noindent\begin{minipage}[l]{\textwidth}
\bfseries
\huge Summary  
\bigskip

\normalsize
\mdseries


Saliency Prediction aims to predict human attention distribution given
an RGB image. More attractive or more salient regions receive higher
predicted scores in the result saliency map, demonstrating more human eye
focus. Most of the recent state-of-the-art methods achieve the goal based
on abstract and deep image feature representations, taking advantage of
traditional CNNs. However, the traditional convolutional structure could
not capture the global feature of the image input well due to its small kernel
size and convolutional computation process. Also, simply building a decoder
net upon the image representation from the encoder does not consider the
high-level factors which might influence the saliency prediction, e.g., objects,
color, light, etc. Most of the methods learn the saliency task without some
reasonable causes closely related to human perception.\\

In this thesis, we first do some investigation on human gaze control in the real
world, where we summarize the types of human gaze control and the factors
affecting where people look. We find evidence in the cognitive science literature that the global view and objects of the scene may play a significant role
in human eye perception. Inspired by this, we design a transformer-based
network that includes unified learning of semantic segmentation to address
the problems we identify. The transformer could introduce global features of
the input images, whose bi-directional self-attention calculation further simulates the human visual system since people tend to acquire a global view of
the scene before determining which areas are salient in their brains. The main
structure of our network makes use of classical bottom-up multi-level feature
fusion, combining the global cue from Transformer and local cues from traditional CNN. Inspired by human perception when looking at a picture, we
introduce another subnet for semantic segmentation. This simulates the human eye perception where recognizing objects in the picture highly correlates
to distributing attention to it. Due to the lack of segmentation annotations
in the public benchmarks and possibly making our multi-task methods easier
to use on other saliency datasets in practice, we employ the PASCAL VOC
2012 semantic segmentation dataset as the training and learning target. In addition, simply building multiple different decoders upon the shared encoder
for multiple tasks could not improve the main task, saliency prediction, very
effectively. We design a multi-task attention module to enhance the interaction between the multiple tasks to address the problem. Our ablation and
visualization experiments show the effectiveness of the proposed ideas. Our
model achieves competitive performance compared to other state-of-the-art
methods in both quantitative and qualitative experiments.

\hspace{0.65 cm}  

\end{minipage}
\vspace*{\fill}
\newpage